\documentclass[journal, transmag]{IEEEtran}

\usepackage[colorlinks,linkcolor=red,anchorcolor=blue,citecolor=green]{hyperref} 
\usepackage{upgreek}
\usepackage{multirow}
\usepackage{float}

\usepackage{cite}

\usepackage{multicol,multicap}
\ifCLASSINFOpdf
   \usepackage[pdftex]{graphicx}
\else
   \usepackage[dvips]{graphicx}
\fi
\usepackage{amsmath}
\usepackage{amssymb}
\usepackage{bbm}
\usepackage{bm}
\usepackage{amsfonts}
\usepackage{array}
\ifCLASSOPTIONcompsoc
  \usepackage[caption=false,font=normalsize,labelfont=sf,textfont=sf]{subfig}
\else
  \usepackage[caption=false,font=footnotesize]{subfig}
\fi
\usepackage{threeparttable} 
\usepackage{booktabs}

\usepackage{fixltx2e}

\usepackage{stfloats}

\ifCLASSOPTIONcaptionsoff
  \usepackage[nomarkers]{endfloat}
 \let\MYoriglatexcaption\caption
 \renewcommand{\caption}[2][\relax]{\MYoriglatexcaption[#2]{#2}}
\fi

\usepackage{url}
\usepackage{color}
\usepackage{balance}
\definecolor{red}{RGB}{238, 44, 40}
\definecolor{blue}{RGB}{40, 40, 238}

\DeclareMathOperator{\st}{s.t.}

\usepackage{caption}
\begin{document}
%
\title{Bridging the Domain Gap in Satellite Pose Estimation: 
a Self-Training Approach based on Geometrical Constraints}
%
%
%

\author{Zi Wang, Minglin Chen, Yulan Guo,~\IEEEmembership{Senior Member,~IEEE,} Zhang Li,~\IEEEmembership{Member,~IEEE,} and Qifeng Yu

\thanks{Manuscript received ; revised . 
    (Corresponding author: Yulan Guo and Zhang Li.)}

\thanks{Zi Wang, Zhang Li and Qifeng Yu are with the
        College of Aerospace Science and Engineering, 
        National University of Defense Technology, Changsha 410073, China, and
        Hunan Provincial Key Laboratory of Image Measurement and Vision Navigation, Changsha 410073, China 
        (E-mail:wangzi16@nudt.edu.cn, \{zhangli\_nudt, qifeng\_yu1958\}@163.com).
    }
\thanks{
        Minglin Chen and Yulan Guo are with the School of Electronics and Communication Engineering,
	Sun Yat-Sen University, Shenzhen 518000, China
	(E-mail: chenmlin8@mail2.sysu.edu.cn, guoyulan@mail.sysu.edu.cn).
	Yulan Guo is also with the College of Electronic Science and Technology,
	National University of Defense Technology, Changsha 410073, China.
	This work was partially done when Zi Wang was a Ph.D. visiting student in SYSU.
}
}

%
%

\markboth{Journal of \LaTeX\ Class Files,~Vol.~14, No.~8, August~2015}%
{Shell \MakeLowercase{\textit{et al.}}: Bare Demo of IEEEtran.cls for IEEE Journals}

\maketitle
\begin{abstract}
    Recently, unsupervised domain adaptation in satellite pose estimation has gained increasing attention,
	aiming at alleviating the annotation cost for training deep models.
    To this end, we propose a self-training framework based on the domain-agnostic \textit{geometrical constraints}.
    Specifically, we train a neural network to predict the 2D keypoints of a 
	satellite and then use PnP to estimate the pose.
    The poses of target samples are regarded as latent variables to formulate the task as a minimization problem.
    Furthermore, we leverage fine-grained segmentation to tackle the information loss 
	issue caused by abstracting the satellite as sparse keypoints.
    Finally, we iteratively solve the minimization problem in two steps:
	pseudo-label generation and network training.
    Experimental results show that our method adapts well to the target domain.
    Moreover, our method won the 1st place on the \texttt{sunlamp} task of 
	the second international Satellite Pose Estimation Competition.    
\end{abstract}

\section{INTRODUCTION}
\label{sec:introduction}
\IEEEPARstart{E}stimating the pose of an uncooperative spacecraft is crucial in numerous space missions, such as debris removal~\cite{Forshaw2016RemoveDEBRISAI}, 
on-orbit servicing~\cite{FloresAbad2014ARO}, and assets refueling~\cite{Malyh2022ABR}.
In the last two decades, numerous relative navigation systems have been proposed based on LiDARs~\cite{Opromolla2019UncooperativeSR,Opromolla2015AM3} 
and cameras~\cite{Naasz2009TheHS,Liu2014RelativePE,Meng2018SatellitePE,Zhang2018VisionBasedPE,Davis2019ProximityOA,kisantal2020satellite}.
Active sensors, including radars and LiDARs, requires larger mass and higher power consumption compared to visual sensors.
Moreover, a stereo system requires a large baseline and relies on robust feature matching to obtain depth information.
Therefore, monocular vision based navigation systems have gained increasing attention from the academical and industrial fields.

Traditionally, monocular satellite pose estimation approaches require hand-crafted feature extraction~\cite{Naasz2009TheHS,Liu2014RelativePE,Meng2018SatellitePE,Zhang2018VisionBasedPE},
which limits their performance on challenging environments, such as occlusions, harsh lighting conditions, reflective materials, and complex structures.
Recently, deep learning based methods have achieved success in satellite pose estimation~\cite{kisantal2020satellite}, 
thanks to the powerful feature representation ability of deep models.
However, training neural networks requires large-scale datasets, while collecting real images of spacecrafts and annotating their 6-DoF poses are 
time-consuming, notoriously laborious, and difficult.
Therefore, recent deep models are trained and evaluated using synthetic images~\cite{chen2019satellite,AAS2019park,Sharma2020NeuralNP,kisantal2020satellite,wang2022revisiting}.
Due to the inherent discrepancy between real and synthetic images, 
the deep models that are fully-supervised by synthetic images usually show deteriorated performance when being deployed in real scenarios. 
This issue is revealed by the large gap between scores on the real and synthetic leaderboards\footnote{\url{https://kelvins.esa.int/satellite-pose-estimation-challenge/leaderboard/leaderboard}}.

In computer vision, unsupervised domain adaptation (UDA)~\cite{hoffman2018cycada,Yaroslav2016Domain,Chen2018DomainAF,Tsai2018LearningTA,Vu2019ADVENTAE,Deng2021UnbiasedMT} 
is adopted in scenarios where the labels of real samples are scarce, 
by training deep models using labeled synthetic images and unlabeled real images. 
Motivated by these practices, Park~\textit{et al.}~\cite{park2021speed+} created the next generation spacecraft pose estimation dataset (SPEED+) 
with focus on the synthetic-to-real domain gap in satellite pose estimation.
Moreover, based on the SPEED+ dataset, the Advanced Concepts Team (ACT) of the European Space Agency (ESA) and  the Space Rendezvous Laboratory (SLAB) at Stanford University 
co-organized the second international Satellite Pose Estimation Competition (SPEC2021)\footnote{\url{https://kelvins.esa.int/pose-estimation-2021/}} to boost the research of bridging the domain gap.

\begin{figure*}[!t]
	\centering
	 \subfloat[\texttt{synthetic}]{
	   \begin{minipage}[b]{0.33\linewidth}
		  \includegraphics[width=\textwidth]{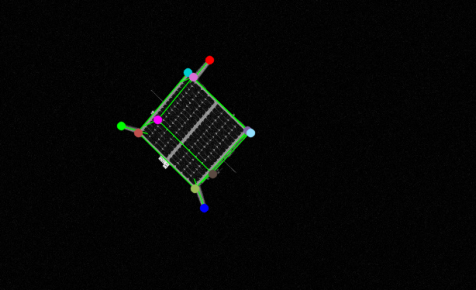} \\\vspace*{-9pt}

		  \includegraphics[width=\textwidth]{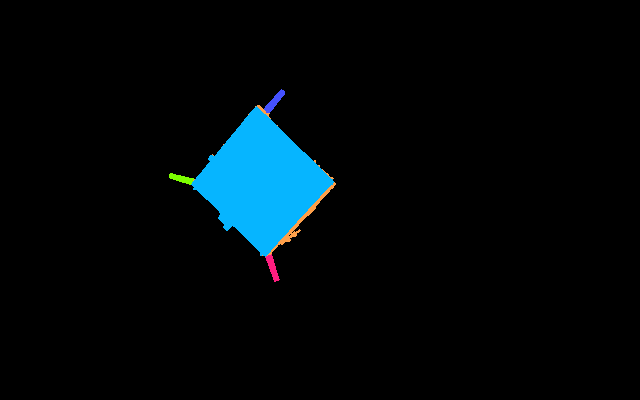}
	   \end{minipage}
	}
	\subfloat[\texttt{lightbox}]{
	   \begin{minipage}[b]{0.33\linewidth}
		\includegraphics[width=\textwidth]{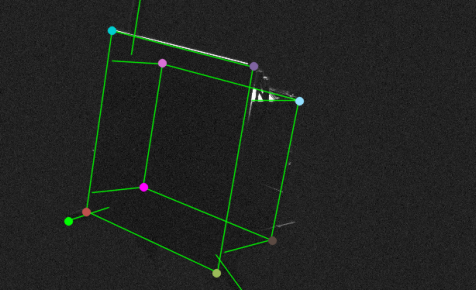} \\ \vspace*{-9pt}

		\includegraphics[width=\textwidth]{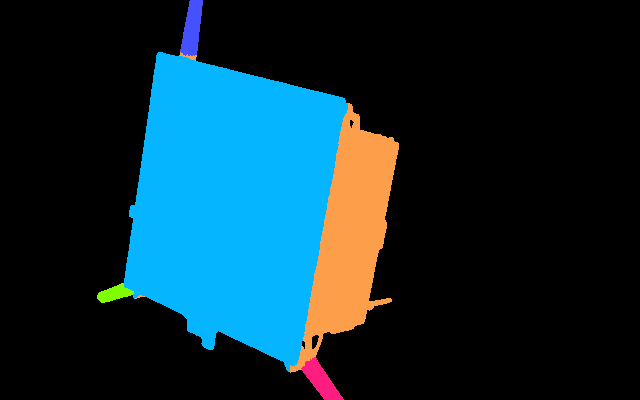}
	   \end{minipage}
	}
	\subfloat[\texttt{sunlamp}]{
	   \begin{minipage}[b]{0.33\linewidth}
		\includegraphics[width=\textwidth]{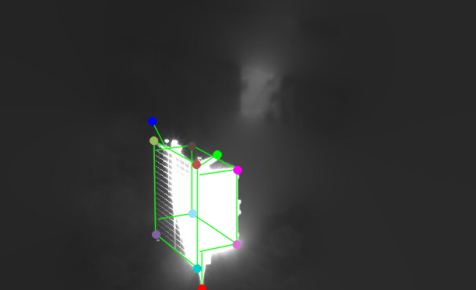} \\\vspace*{-9pt}

		\includegraphics[width=\textwidth]{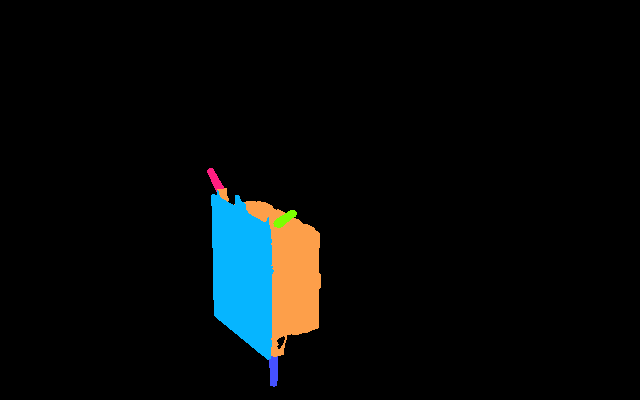}
	   \end{minipage}
	}
	\caption{
	The characteristic of the UDA task in satellite pose estimation. 
    The satellite structures are the same in different domains with different illumination conditions.
	The samples come from three distinct domains, including \texttt{synthetic} (well illuminated), 
	\texttt{lightbox} (low contrast, highly noisy), and \texttt{sunlamp} (specular reflection).
	Top: the original images from~\cite{park2021speed+} with the 2D keypoints and the reprojected wireframe model. 
	Bottom: the fine-grained masks. 
	Note that, the visualization results on the \texttt{synthetic} image is obtained using the ground-truth pose,
	while the poses used on the \texttt{lightbox} and \texttt{sunlamp} images are predicted by our network.
	}
	\label{fig:labels}
\end{figure*}

Different to common UDA tasks in computer vision, a calibrated camera is used to measure the pose of the same satellite under various environments in space missions.
For UDA in satellite pose estimation in SPEC2021, 
only the environmental settings are different across domains while the satellite structure and the camera parameters are the same (as shown in Fig.~\ref{fig:labels}).
Therefore, given a pose, the locations of keypoints and the satellite masks are the same in different domains, 
which are referred to as the domain-agnostic \textit{geometrical constraints}.
Besides, due to intense imaging noise, challenging illumination variations, and diverse poses, 
this task poses additional challenges compared with common UDA problems~\cite{Yaroslav2016Domain,Tsai2018LearningTA,Chen2018DomainAF}.

Several UDA approaches~\cite{zou2018unsupervised, li2019bidirectional,long2016unsupervised,Vu2019ADVENTAE,Deng2021UnbiasedMT,wang2021occlusion} 
have been proposed to explore self-training paradigms to improve performance on the target domain.
Nonetheless, these methods are not specifically designed for UDA in satellite pose estimation, since the domain-agnostic \textit{geometrical constraints} are not fully explored.
Meanwhile, previous satellite pose estimation approaches~\cite{chen2019satellite, wang2022revisiting} represent a satellite as a set of 2D keypoints, 
and then estimate the satellite pose using the perspective-n-point (PnP) algorithms~\cite{gao2003complete}.
However, as shown in Fig.~\ref{fig:labels}, sparse keypoints stand for only semantic parts of the satellite.
Such sparse representation leads to a significant loss of information, which hampers knowledge transfer across domains.

To tackle the above problems, we formulate UDA of satellite pose estimation as a minimization problem under a self-training framework. 
First, we formulate the \textit{geometrical constraints} as a projection function, 
which maps the predefined 3D keypoints onto the source and target images using the same camera parameters.
Based on the projection function, we propose a basic self-training framework by taking the poses of target samples as latent variables, 
which are jointly optimized with the network parameters.
Second, we leverage fine-grained segmentation to extend the basic framework.
Specifically, we enhance the \textit{geometrical constraints} with a rendering function.
Similar to the projection function, the rendering function maps the 3D mesh of the satellite to fine-grained masks, 
using the same camera parameters in different domains.
Therefore, we take fine-grained segmentation as an auxiliary task of keypoints regression.
Furthermore, as the masks provide dense descriptions with structural information,
we perform adversarial training by aligning the predicted masks of the source and target samples.
Finally, we iteratively optimize the network parameters and generate pseudo labels to solve the minimization problem.
Experimental results demonstrate the effectiveness of our framework.
Moreover, our method won the 1st and the 3rd place on two leaderboards of the second international Satellite Pose Estimation 
Competition\footnote{\url{https://kelvins.esa.int/pose-estimation-2021/} team: lava1302}, respectively.

Our contributions can be summarized as follows:
(1) We explore the domain-agnostic \textit{geometrical constraints} to propose a self-training framework for UDA in satellite pose estimation.
(2) We leverage fine-grained masks to address the information loss problem caused by abstracting the satellite as sparse keypoints.
(3) Our method significantly improves the accuracy of satellite pose estimation without using real annotations.

\section{RELATED WORK}
\textbf{Object pose estimation} aims at recovering the 3D position and 3D rotation of an object in the camera-centered coordinate system.
Traditional approaches~\cite{Grundmann2011AGM,Brachmann2016UncertaintyDriven6P} rely on local features, 
suffering from texture-less objects and background clutter.
Recently, CNN-based methods have dominated most object pose estimation tasks.
Numerous approaches~\cite{Tekin2018RealTimeSS, Hu2019SegmentationDriven6O, zakharov2019dpod,
li2019cdpn, Hodan2020EPOSE6, peng2020pvnet} have been proposed to estimate poses using putative 2D-3D correspondences 
 and the PnP algorithms.
To achieve better efficiency, several methods~\cite{Kehl2017SSD6DMR, Hu2020SingleStage6O, Labbe2020CosyPoseCM, Wang2021GDRNetGD} 
are introduced to directly regress poses from monocular images.
Other methods~\cite{sundermeyer2020augmented, sundermeyer2020multi,Wen2020EdgeEI} 
learn the latent representations of rotation and recover poses by exploring the image retrieval paradigms.
Since these methods focus on household objects in indoor scenarios~\cite{hinterstoisser2012model, xiang2017posecnn},
they face significant challenges caused by wide-range depth variations and illumination changes in outer space~\cite{hu2021wide}.

\textbf{Satellite pose estimation} is a special case of object pose estimation.
Spacecraft Pose Network (SPN)~\cite{Sharma2020NeuralNP} is the first deep learning-based approach for satellite pose estimation. 
Specifically, the 3D rotation is recovered by discretizing the viewpoint spaces into bins, 
and then the 3D translation is estimated using the \textit{geometrical constraints}.
In other top-performing approaches, 
the pose estimation problem is formulated as a task of localizing semantic keypoints on the convex areas of a satellite by taking various representations, 
such as heatmap \cite{chen2019satellite}, vector \cite{AAS2019park}, and set \cite{wang2022revisiting}. 
These methods crop the satellite from the input images using a well-trained object detector to address scale variations. 
To achieve better efficiency, Hu \textit{et al}~\cite{hu2021wide} handle the scale problem in a single-stage way by introducing a sampling strategy.
However, these methods are trained and tested on synthetic data. 
They usually undergo significant performance degradation when being applied to real images due to the domain gap~\cite{Wilson2020ASO}.

\textbf{Unsupervised domain adaptation} aims at addressing the domain mismatch problem.
It is a promising direction to circumvent the laborious and time-consuming procedures of data annotation.
Several UDA paradigms have been studied for different vision tasks.
Adversarial learning aligns both domains with a discriminator. 
The alignment can be achieved at image-level~\cite{hoffman2018cycada}, 
feature-level~\cite{Yaroslav2016Domain, Chen2018DomainAF}, and output-level~\cite{Tsai2018LearningTA}.
Self-training methods are introduced to utilize target samples to train the model by generating pseudo labels~\cite{zou2018unsupervised, li2019bidirectional}, 
minimizing the entropy loss~\cite{long2016unsupervised, Vu2019ADVENTAE}, or employing the teacher-student framework~\cite{Deng2021UnbiasedMT,wang2021occlusion}.
Nonetheless, it is challenging to adopt these approaches in UDA of satellite pose estimation, 
which is different to the common UDA tasks in computer vision.

\section{METHOD}\label{sec:method}
In this section, we first introduce the UDA task of satellite pose estimation (Sec.~\ref{sec:problem}) and 
the PnP-based solution to monocular satellite pose estimation (Sec.~\ref{sec:pnp}).
Then, we formulate the task as a minimization problem in a basic self-training framework (Sec.~\ref{sec:basic}), 
which is extended by leveraging fine-grained segmentation (Sec.~\ref{sec:extend}).
We present the solution to the minimization problem (Sec.~\ref{sec:solution}).
Finally, we give a mathematical proof of the 
\textit{geometrical constraints} (Sec.~\ref{sec:discussion}).
The overview of our method is shown in Fig.~\ref{fig:overview}.

\subsection{Problem Formulation}\label{sec:problem}
In UDA of satellite pose estimation, the satellite structure and the camera parameters are  the same in the source and target domains.
The intrinsic matrix of the camera is denoted by $\mathbf{K}$.
Then, we are given $N_s$ source images $\mathcal{X}_s=\{\mathbf{I}_s^i\}_{i=1}^{N_s}$ 
with 6-DoF pose annotations $\mathcal{T}_s=\{[\mathbf{R}_s^i|\bm{t}_s^i] \}_{i=1}^{N_s}$ 
and $N_t$ unlabeled target images $\mathcal{X}_t=\{ \mathbf{I}_t^j \}_{j=1}^{N_t}$.

We assume that the poses in the source and target domains are sampled from the same distribution.
\begin{equation}\label{eq:pose_eq}
    P(\mathbf{R}_s, \bm{t}_s) = P(\mathbf{R}_t, \bm{t}_t)
\end{equation}
However, given a pose consisting of $\mathbf{R}$ and $\bm{t}$, the source and target images are sampled from different distributions,
\begin{equation}
	P(\mathbf{I}_s|\mathbf{R}_s, \bm{t}_s) \neq P(\mathbf{I}_t|\mathbf{R}_s, \bm{t}_s)
\end{equation}
It reveals that the source and target samples are sampled from different joint distributions,
\begin{align}~\label{eq:iid}
	&P(\mathbf{I}_s, \mathbf{R}_s, \bm{t}_s) = P(\mathbf{I}_s|\mathbf{R}_s, \bm{t}_s) P(\mathbf{R}_s, \bm{t}_s)\neq\\
	&P(\mathbf{I}_t, \mathbf{R}_t, \bm{t}_t) = P(\mathbf{I}_s|\mathbf{R}_t, \bm{t}_t) P(\mathbf{R}_t, \bm{t}_t) \nonumber
\end{align}
Equation~\ref{eq:iid} shows that the \textit{independent and identically distributed (i.i.d.)} assumption is violated, 
which is referred to as the domain shifts in satellite pose estimation.
In this paper, we aim to train a neural network $\mathcal{G}(\mathbf{I})\mapsto [\mathbf{R}|\bm{t}]$,
that reduces the distribution shifts across domains.

\subsection{PnP-based Solution}\label{sec:pnp}
As suggested by Kisantal~\textit{et al.}~\cite{kisantal2020satellite},
the PnP-based methods significantly outperform direct regression in monocular satellite pose estimation.
Therefore, we follow the PnP based~\cite{chen2019satellite,wang2022revisiting} method to estimate the satellite poses.

We assume the texture-less 3D mesh $\mathcal{M}$ of the satellite is given.
Hence, we select $N_p$ 3D landmarks $\mathcal{P} = \{\bm{P}^k\}_{k=1}^{N_p}$ on the mesh surface.  
Given the camera parameters and a pose, 
the 3D landmarks are reprojected onto 2D keypoints $\{\bm{p}^{k}\}_{k=1}^{N_p}$, 
which are represented as a heatmap $\mathbf{H}$.
We construct a neural network consisting of a backbone $\mathcal{G}_f$ and a heatmap head $\mathcal{G}_h$ and 
train the network using source samples in a fully-supervised way,
\begin{equation}\label{eq:source}
	\min_{\mathcal{G}_{f,h}} \frac{1}{N_s}\sum_{i=1}^{N_s} \mathcal{L}_h(\hat{\mathbf{H}}_s^i, \mathbf{H}_s^i)
\end{equation}
where $\hat{\mathbf{H}}_s^i = \mathcal{G}_f(\mathcal{G}_h(\mathbf{I}_s))$ is the predicted heatmap, 
$\mathbf{H}_s^i$ is the ground truth.
For heatmap regression, 
$\mathcal{L}_h$ is adopted as the adaptive wing loss~\cite{Wang2019AdaptiveWL}
with the default parameter setting.
During the inference stage, the predicted heatmap $\hat{\mathbf{H}}$ is decoded into 2D keypoints, 
which are used to build putative 2D-3D correspondences $\{(\hat{\bm{p}}^k, \bm{P}^k)\}_{k=1}^{N_p}$.
Finally, the pose is estimated by solving the PnP problem~\cite{wang2022revisiting},
\begin{equation}\label{eq:pnp}
	\hat{\mathbf{R}}, \hat{\bm{t}} = \min_{\mathbf{R},\bm{t}}
	\sum_{k=1}^{N_p} \phi(\|\lambda_k \hat{\bm{p}}^k- \mathbf{K}(\mathbf{R}\bm{P}^k + \bm{t})\|_2)
\end{equation}
where $\lambda_k $ is the depth of landmark $\bm{P}^k$ and $\phi(\cdot)$ is the Huber loss.

Due to the domain shifts, the neural network that trained on the source domain usually 
suffers from performance degradation when being applied on target images.
To tackle this issue, we train the network using labeled source samples and unlabeled target samples (as described in Sec.~\ref{sec:basic}).  

\subsection{Basic Self-Training Framework}\label{sec:basic}
To fully exploit unlabeled target samples, we propose a self-training framework by leveraging the \textit{geometrical constraints}.
Specifically, we define the function that projects 3D landmarks onto the 2D heatmap as,
\begin{equation}\label{eq:f_func}
	\mathbf{H}=\mathcal{F}(\mathbf{R}, \bm{t}, \mathcal{P}, \mathbf{K})
\end{equation}
which provides the same \textit{geometrical constraints} in the source and target domains,
since the source and target samples share the same $\mathcal{P}$ and $\mathbf{K}$.

For source samples, we supervise the network using Eq.~(\ref{eq:source}).
The ground-truth heatmap $\mathbf{H}_s$ is obtained by projecting landmarks $\mathcal{P}$ 
using the function $\mathcal{F}$ with the pose annotation $[\mathbf{R}_s| \bm{t}_s]$.
For unlabeled target samples, we take each target pose as a latent variable, 
which is sent to the function $\mathcal{F}$ to obtain the pseudo heatmap $\tilde{\mathbf{H}}_t$.
Therefore, we can unify Eq.~(\ref{eq:source}) and (\ref{eq:pnp}) in an objective function to exploit unlabeled target images.
Specifically, the task is formulated as a minimization perform by simultaneously optimizing the network parameters and the target poses,
\begin{align}\label{eq:self1}
	&\min_{\mathcal{G}_{f,h}, \mathcal{T}_t} 
	\frac{1}{N_s}\sum_{i=1}^{N_s} \mathcal{L}_h(\hat{\mathbf{H}}_s^i, \mathbf{H}_s^i) + 
	\frac{1}{N_t}\sum_{j=1}^{N_t}
		\mathcal{L}_h(\hat{\mathbf{H}}_t^j, \tilde{\mathbf{H}}_t^j) \\
	&\begin{array}{r@{\;}l}
		\st\quad  &\tilde{\mathbf{H}}_t^j  = \mathcal{F}(\mathbf{R}_t^j, \bm{t}_t^j, \mathcal{P}, \mathbf{K}),\nonumber\\
		&\mathcal{T}_t^i = [\mathbf{R}_t^j|\bm{t}_t^j] \in \mathbb{SE}_3 \cup \{\mathbf{0}\}, \;
		j = 1, 2, \dots, N_t 
	\end{array}
\end{align}
where the zero set $\{\mathbf{0}\}$ is introduced to tackle the issues of less confident predictions on target samples,
which are ignored during the network training by assigning their poses to $\mathbf{0}$.
The criterion to select confident predictions is presented in Sec.~\ref{sec:solution}.

Equation~\ref{eq:self1} can be solved iteratively.
During optimization, the network $\mathcal{G}_{f,h}$ is encouraged to predict 
accurate heatmaps on source and target images.
Meanwhile, the target poses should converge to the ground-truth.
However, the heatmap-based representation of sparse keypoints introduces severe information loss.
Consequently, the optimization process is prone to converging at local optima, leading to deteriorated performance.

\begin{figure*}[!t]
	\includegraphics[width=\textwidth]{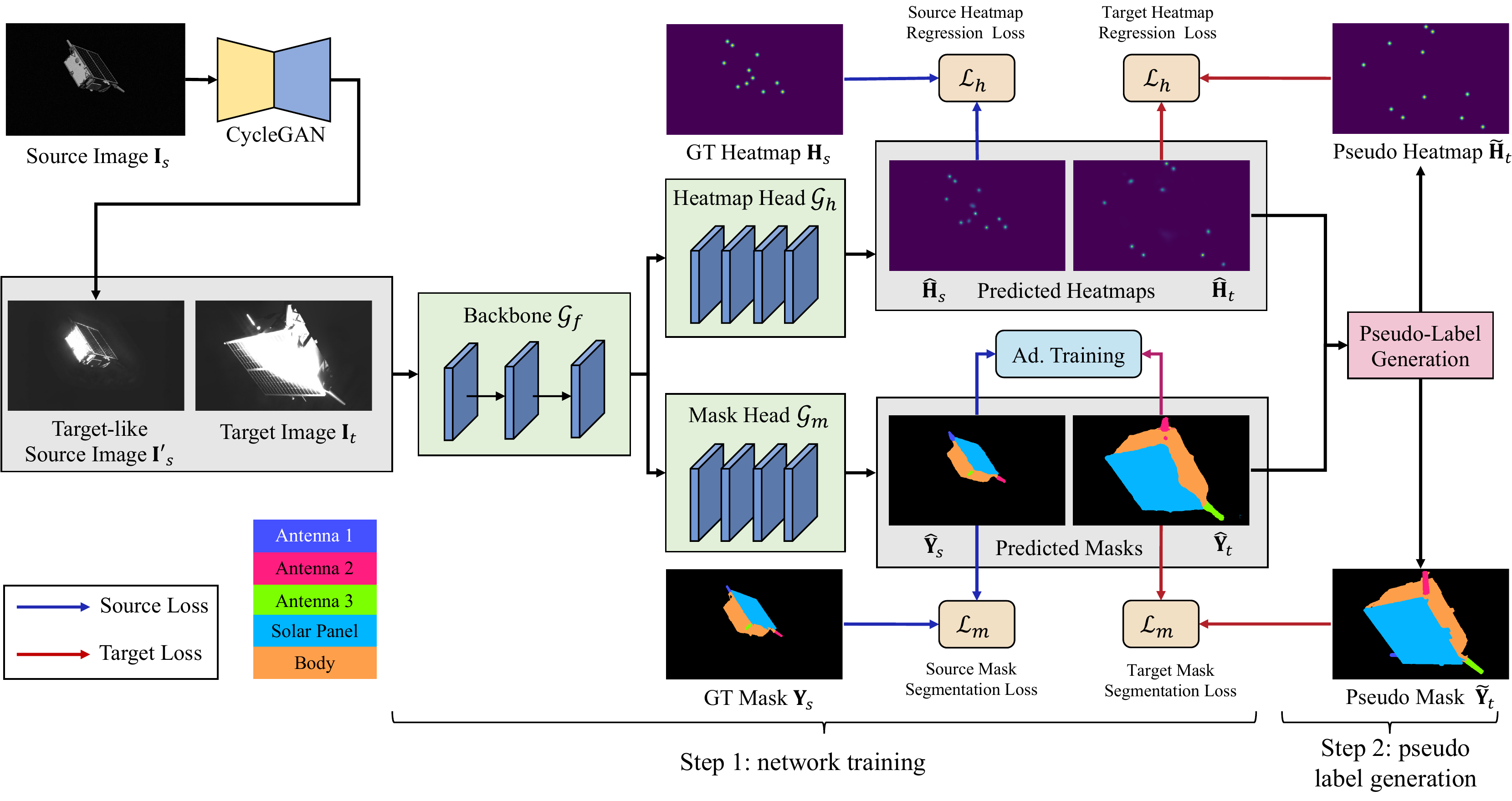}
\caption{Overview of our self-training framework.
The satellite is represented as a set of sparse keypoints and dense fine-grained masks.
The source image $\mathbf{I}_s$ is first transformed into a target-like source image $\mathbf{I}'_s$ using a well-trained CycleGAN~\cite{Zhu2017UnpairedIT}. 
The neural network consists of a backbone $\mathcal{G}_f$, a heatmap head $\mathcal{G}_h$, 
and a mask head $\mathcal{G}_m$. The optimization comprises two iterative steps.
In the first step, pseudo mask $\tilde{\mathbf{Y}}_t$ and pseudo heatmap $\tilde{\mathbf{H}}_t$ are fixed, 
and the network $\mathcal{G}_{f,h,g}$ are trained using multi-task losses.
In the second step, the parameters of the neural network are fixed, 
and the predicted mask $\hat{\mathbf{Y}}_t$ and heatmap $\hat{\mathbf{H}}_t$ are used to generate pseudo labels.
(Best viewed in color.)}
\label{fig:overview} 
\end{figure*}

\subsection{Extended Framework with Multi-task Learning}\label{sec:extend}
To tackle the above problem, we note that the fine-grained masks in Fig.~\ref{fig:labels} provide
rich dense descriptions with domain-agnostic structural context.
Therefore, we apply segmentation as an auxiliary task of heatmap regression to improve pose estimation performance.
In the following paragraphs, we first perform output-level alignment~\cite{Tsai2018LearningTA} using adversarial training, 
and then extend the minimization problem in Eq.~(\ref{eq:self1}) with the auxiliary task. 

As shown in Fig.~\ref{fig:overview}, 
we extend the basic network with a mask head $\mathcal{G}_m$ to predict the fine-grained mask 
$\hat{\mathbf{Y}} = \mathcal{G}_m(\mathcal{G}_f(\mathbf{I}))$ for each image.
These masks contain domain-agnostic information, 
which can be used to align the source and target samples by adversarial training.
Hence, we build a discriminator $\mathcal{G}_d$ to predict the domain label of each sample.
The discriminator $\mathcal{G}_d$ receives the detached mask predictions from both domains and is trained using the BCE loss,
\begin{align}
	\mathcal{L}_{d}(\hat{\mathbf{Y}}_t, \hat{\mathbf{Y}}_s) = -\frac{1}{HW}\sum_{i=1}^H\sum_{j=1}^W &
	\log(\mathcal{G}_d(\hat{\mathbf{Y}}_s)(i, j, 1)) + \nonumber\\
	&\log(\mathcal{G}_d(\hat{\mathbf{Y}}_t)(i, j, 0))
\end{align}
where $\hat{\mathbf{Y}}_s$ and $\hat{\mathbf{Y}}_t$ are the predicted masks of the source and target samples, respectively.
To bridge the domain gap, 
we expect the mask predicted by the network $\mathcal{G}_{f,m}$ to fool the discriminator.
Therefore, we perform an adversarial loss on the predicted masks of the target images and backpropagate the gradient.
\begin{equation}
  \mathcal{L}_{adv}(\hat{\mathbf{Y}}_t) = -\frac{1}{HW}\sum_{i=1}^H\sum_{j=1}^W\log(\mathcal{G}_d(\hat{\mathbf{Y}}_t)(i, j, 1))
\end{equation}

Furthermore, we introduce a rendering function $\mathcal{R}$ to enhance the \textit{geometrical constraints}.
Given the pose $[\mathbf{R}|\bm{t}]$, the mask $\mathbf{Y}$ can be obtained through the rendering function, 
\begin{equation}
	\mathbf{Y} = \mathcal{R}(\mathbf{R}, \bm{t}, \mathcal{M}, \mathbf{K})
\end{equation}
Similar to the projection function $\mathcal{F}$,
the rendering function $\mathcal{R}$ also provides the same \textit{geometrical constraints} on source and target samples. 
This is because the 3D mesh of the satellite and the camera parameters are the same in different domains.
For source samples, the network $\mathcal{G}_{f,m}$ is supervised using the ground-truth masks.
For unlabeled target images, the pseudo masks are obtained by sending the latent variables in Eq.~(\ref{eq:self1}), 
\textit{i.e.}, the target poses, to the rendering function $\mathcal{R}$.  
Therefore, the minimization problem depicted in Eq.~(\ref{eq:self1}) is extended  with the tasks of segmentation and adversarial training,
\begin{align}\label{eq:self2}
	&\begin{array}{*3{>{\displaystyle}l}p{0.1cm}}
    &\min_{\mathcal{G}_{f,h,m}, \mathcal{T}_t} \max_{\mathcal{G}_d}
	\frac{1}{N_s}\sum_{i=1}^{N_s} \left[ \mathcal{L}_h(\hat{\mathbf{H}}_s^i, \mathbf{H}_s^i) + 
	\lambda_m \mathcal{L}_m(\hat{\mathbf{Y}}_s^i, \mathbf{Y}_s^i)\right] + \\
    & \frac{1}{N_t}\sum_{j=1}^{N_t} \left[
	   \mathcal{L}_h(\hat{\mathbf{H}}_t^j, \tilde{\mathbf{H}}_t^j) + 
    	\lambda_m \mathcal{L}_m(\hat{\mathbf{Y}}_t^j, \tilde{\mathbf{Y}}_t^j) +
		\lambda_a \mathcal{L}_{adv}({\hat{\mathbf{Y}}_t^j}) 
	\right] \\
	\end{array}\\
	&\begin{array}{r@{\;}l}
		\st\quad  
        &\tilde{\mathbf{H}}_t^j  = \mathcal{F}(\mathbf{R}_t^j, \bm{t}_t^j, \mathcal{P}, \mathbf{K}),\;
		\tilde{\mathbf{Y}}_t^j  = \mathcal{R}(\mathbf{R}_t^j, \bm{t}_t^j, \mathcal{M}, \mathbf{K}),\nonumber\\
        &\mathcal{T}_t^i = [\mathbf{R}_t^j|\bm{t}_t^j] \in \mathbb{SE}_3 \cup \{\mathbf{0}\} \vspace{1ex},\;
		j = 1, 2, \dots, N_t \nonumber\\
	\end{array}
\end{align}
where $\lambda_m$ and $\lambda_a$ are the weights for the mask loss and the adversarial loss, respectively.
$\mathcal{L}_{m}$ is the cross-entropy loss for segmentation.
We refer to $\tilde{\mathbf{Y}}$ as the pseudo mask.
Again, by optimizing the loss in Eq.~(\ref{eq:self2}), 
the network $\mathcal{G}_{f,h,m}$ should perform well on the target domain, 
while the pseudo labels, including pseudo heatmaps, masks, and poses, should approximate the ground truth.

\subsection{Iterative Optimization and Pseudo Label Generation}\label{sec:solution}
We observe that the variables in Eq.~(\ref{eq:self2}) can be divided into two classes: 
the network parameters and the poses of target samples.
Following \cite{zou2018unsupervised}, we adopt iterative procedures to optimize Eq.~(\ref{eq:self2}):
	\setlength{\topsep}{0pt}
	\begin{enumerate}
		\setlength{\itemsep}{-1pt}
		\item Fix the pseudo labels (or initialize $\mathcal{T}_t$ as $\{\mathbf{0}\}$) and train the network $\mathcal{G}_{f,h,m,d}$;
		\item Fix the network, optimize poses $\mathcal{T}_t$,
		 and generate pseudo labels.
	\end{enumerate}
We take the combination of these two steps as one round and take several rounds to optimize Eq.~(\ref{eq:self2}).

In the first step, when the pseudo heatmaps and pseudo masks are fixed, 
the minimization problem in Eq.~(\ref{eq:self1}) is simplified as,
\begin{equation}
	\begin{array}{*3{>{\displaystyle}l}p{0.1cm}}
		&\min_{\mathcal{G}_{f,h,m}} \max_{\mathcal{G}_d}
		\frac{1}{N_s}\sum_{i=1}^{N_s} \left[ \mathcal{L}_h(\hat{\mathbf{H}}_s^i, \mathbf{H}_s^i) + 
		\lambda_m \mathcal{L}_m(\hat{\mathbf{Y}}_s^i, \mathbf{Y}_s^i)\right] + \\
		& \frac{1}{N_t}\sum_{j=1}^{N_t} \left[
			\mathcal{L}_h(\hat{\mathbf{H}}_t^j, \tilde{\mathbf{H}}_t^j) + 
			\lambda_m \mathcal{L}_m(\hat{\mathbf{Y}}_t^j, \tilde{\mathbf{Y}}_t^j) +
			\lambda_a \mathcal{L}_{adv}({\hat{\mathbf{Y}}_t^j}) 
		\right]
	\end{array}
\end{equation}
It is equal to train the neural network using the labeled source samples and the pseudo-labeled target samples.
In the second step, when the network parameters are fixed, 
the minimization problem in Eq.~(\ref{eq:self2}) is simplified as,
\begin{align}
    &\min_{\mathcal{T}_t}
    \frac{1}{N_t}\sum_{j=1}^{N_t} \left[
	   \mathcal{L}_h(\hat{\mathbf{H}}_t^j, \tilde{\mathbf{H}}_t^j) + 
    	\lambda_m \mathcal{L}_m(\hat{\mathbf{Y}}_t^j, \tilde{\mathbf{Y}}_t^j) 
	\right] \\
	&\begin{array}{r@{\;}l}
		\st\quad  
        &\tilde{\mathbf{H}}_t^j  = \mathcal{F}(\mathbf{R}_t^j, \bm{t}_t^j, \mathcal{P}, \mathbf{K}),\;
		\tilde{\mathbf{Y}}_t^j  = \mathcal{R}(\mathbf{R}_t^j, \bm{t}_t^j, \mathcal{M}, \mathbf{K}),\nonumber\\
        &\mathcal{T}_t^i = [\mathbf{R}_t^j|\bm{t}_t^j] \in \mathbb{SE}_3 \cup \{\mathbf{0}\} \vspace{1ex},\;
		j = 1, 2, \dots, N_t \nonumber\\
	\end{array}
\end{align}
It is equal to optimize the target poses to minimize the loss,
by simultaneously aligning the heatmaps and masks generated using functions $\mathcal{F}$ and $\mathcal{R}$ with the ones predicted by the network.
However, for the classification loss $\mathcal{L}_m$, 
there is a trivial solution by ignoring all pseudo heatmaps, \textit{i.e.}, $\mathcal{T}_s=\{\mathbf{0}\}$.
Therefore, we leverage the predicted heatmaps to estimate poses and then generate pseudo labels.

Specifically, we decode the predicted heatmap $\hat{\mathbf{H}}_t$ of the target image into the 2D coordinates of keypoints, 
which are then used to generate putative correspondences $\{(\hat{\bm{p}}_t^k, \bm{P}^k)\}_{k=1}^{N_p}$.
Next, the estimated pose $[\hat{\mathbf{R}}|\hat{\bm{t}}]$ and the number of inliers $N_{in}$ can be obtained by solving Eq.~(\ref{eq:pnp}).
We kindly refer to the readers to our previous work~\cite{wang2022revisiting} for the detail solution to Eq.~(\ref{eq:pnp}).
Furthermore, we take $N_{in}$ as the confidence of the estimated pose and select the estimated pose as a pseudo pose if $N_{in} \geq  N_{th}$.
Otherwise, the pseudo pose is set to $\textbf{0}$ and the corresponding sample is excluded during network retraining.
A smaller $N_{th}$ encourages more estimated poses to be used to generate pseudo labels for model training.
Additionally, for target samples with $N_{in} < N_{th}$, we generate pseudo masks using the approach proposed by~\cite{li2019bidirectional}.

\subsection{Discussion}\label{sec:discussion}
Our self-training framework is built upon the \textit{geometrical constraints}, 
which are provided by the projection function $\mathcal{F}$ and the rendering function $\mathcal{R}$.
We try to provide a mathematical proof of the \textit{geometrical constraints} in this section from the perspective of probability distribution.
Since the functions $\mathcal{F}$ and $\mathcal{R}$ have the same formation, we only present the proof of the function $\mathcal{F}$ in the remainder of this section.

Equation~(\ref{eq:f_func}) has four input parameters, 
including satellite pose $[\mathbf{R}|\bm{t}]$, 3D landmarks $\mathcal{P}$, 
and camera intrinsic matrix $\mathbf{K}$.
Since $\mathcal{P}$ and $\mathbf{K}$ are constant, 
if the poses of the satellite in the source and target domain are equal,
the heatmaps in both domains are the same.
\begin{equation}\label{eq:equal}
\begin{split}
      &[\mathbf{R}_s|\bm{t}_s]=[\mathbf{R}_t|\bm{t}_t] \Rightarrow \\
	 & \mathcal{F}(\mathbf{R}_s, \bm{t}_s, \mathcal{P}, \mathbf{K}) = 
	 \mathcal{F}(\mathbf{R}_t, \bm{t}_t, \mathcal{P}, \mathbf{K}) \Rightarrow \\ 
      & \mathbf{H}_s = \mathbf{H}_t
\end{split}
\end{equation}
Therefore, when conditioned on the satellite pose, the heatmap distributions in source and target domains are the same.
\begin{equation}\label{eq:condition}
    P(\mathbf{H}_s|\mathbf{R}_s, \bm{t}_s) = P(\mathbf{H}_t|\mathbf{R}_t, \bm{t}_t) 
\end{equation}
On the other hand, the heatmap distributions can be marginalized on the satellite poses as following,
\begin{equation}\label{eq:marginalize}
\begin{split}
    P(\mathbf{H}_s) &= \int\int P(\mathbf{H}_s|\mathbf{R}_s, \bm{t}_s) P(\mathbf{R}_s=\mathbf{R}, \bm{t}_s=\bm{t}) \;\mathrm{d}\mathbf{R}\mathrm{d}\bm{t}\\
    P(\mathbf{H}_t) &= \int\int P(\mathbf{H}_t|\mathbf{R}_t, \bm{t}_t) P(\mathbf{R}_t=\mathbf{R}, \bm{t}_t=\bm{t}) \;\mathrm{d}\mathbf{R}\mathrm{d}\bm{t}\\
\end{split}
\end{equation}
By combining Eq.~(\ref{eq:pose_eq}), Eq.~(\ref{eq:condition}), and Eq.~(\ref{eq:marginalize}),
we can get the conclusion that the heatmaps in the source and target domain have the same distribution,
\textit{i.e.},
\begin{equation}
    P(\mathbf{H}_s) = P(\mathbf{H}_t)
\end{equation}
Therefore, the geometrical constraint provided by the projection function $\mathcal{F}$ is domain-agnostic.

What'more, the similar conclusion also holds for the rendering function $\mathcal{R}$, 
\textit{i.e.}, $P(\mathbf{Y}_s) = P(\mathbf{Y}_t)$. 
The fine-grained mask provides dense representation, 
while the keypoints heatmap is sparse and leads to significant information loss.
Hence, we perform output level adaptation to match mask distributions between source and target domain in Sec.~\ref{sec:extend}.

\section{EXPERIMENTAL RESULTS}\label{sec:result}
We present experimental details and results in this section.
We first introduce the dataset and the metrics used in the experiments in Sec.~\ref{sec:dset}.
Then, Sec.~\ref{sec:imp} presents the details of implementation.
Next, the key components of the proposed approach are studied in Sec.~\ref{sec:abl}.
We compare our approach with the state-of-the-art methods in Sec.~\ref{sec:comp}.
Finally, Sec.~\ref{sec:runtime} presents the runtime analysis.

\begin{figure*}[t]
    \centering
    \includegraphics[width=\linewidth]{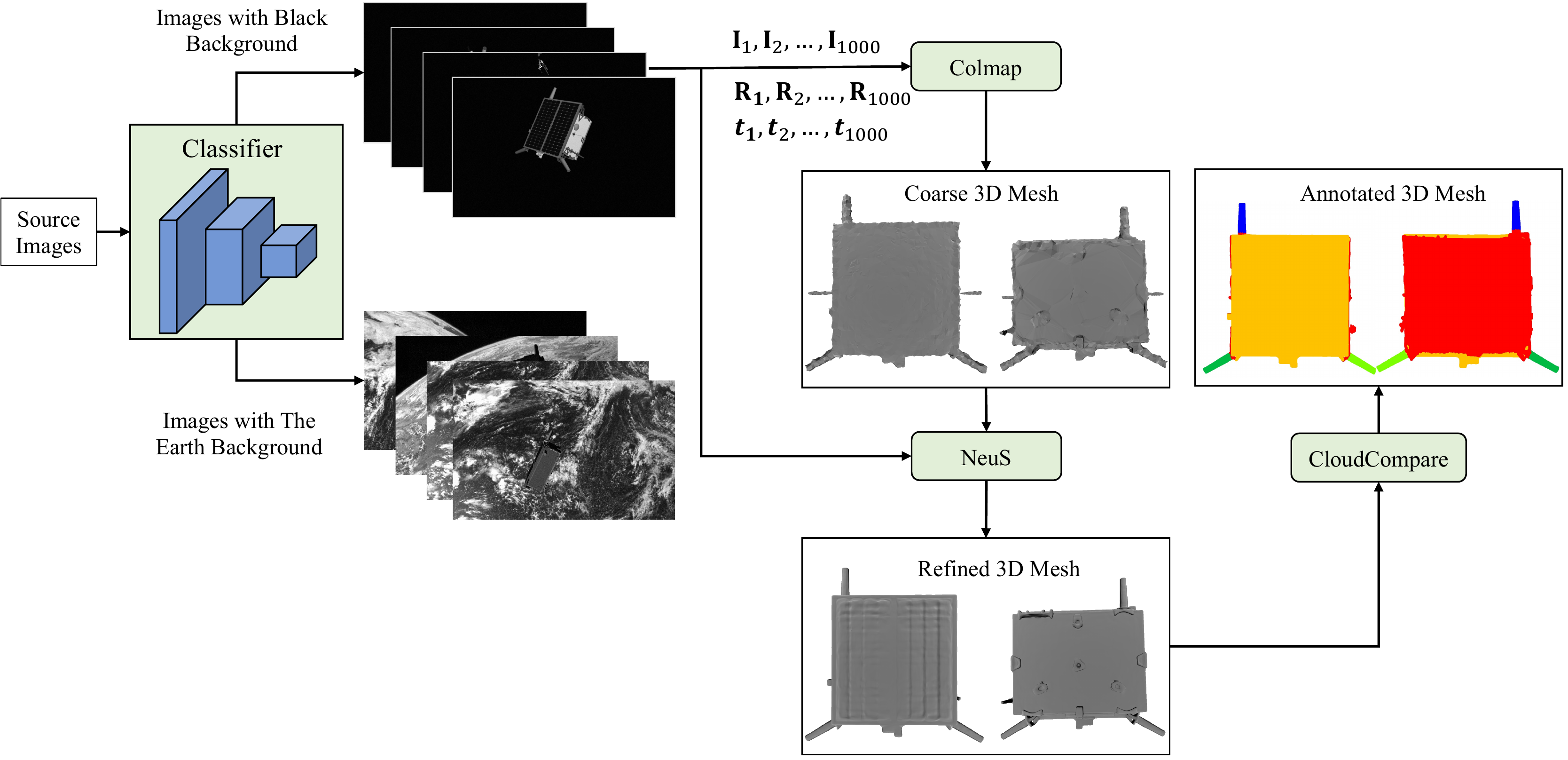}
    \caption{Pipeline of 3D mesh reconstruction and annotation.}\label{fig:recon}
\end{figure*}

\subsection{Dataset and Metrics}\label{sec:dset}
\textbf{Dataset}. 
We conduct experiments on the SPEED+~\cite{park2021speed+} dataset to demonstrate the effectiveness of our method.
The SPEED+~\cite{park2021speed+} dataset comprises images of the Tango spacecraft from the PRISMA~\cite{d2014pose} mission,
consisting of three distinct domains, \textit{i.e.}, \texttt{synthetic}, \texttt{lightbox}, and \texttt{sunlamp}.
Each image has a resolution of $1920 \times 1200$ and contains a single object.
The \texttt{synthetic} domain comprises 59,960 images labeled with poses, 
which are generated using an OpenGL-based stimulator.
The \texttt{lightbox} and \texttt{sunlamp} domains contain 6,740 and 2,791 images of a model of the same spacecraft captured in a robotic simulation environment.
The satellite in the \texttt{lightbox} domain is illuminated by several lightboxes to approximate the diffuse light of Earth,
while the same object in the \texttt{sunlamp} domain is exposed to an arc lamp to simulate the direct sunlight.
Since the annotations for \texttt{lightbox} and \texttt{sunlamp} domains are not released,
there are two UDA tasks with respect to SPEED+, 
including \texttt{synthetic}$\to$\texttt{lightbox} and \texttt{synthetic}$\to$\texttt{sunlamp}.

\textbf{Metrics}.
We adopt the metrics used in SPEC2021.
The rotation error is defined as the angle between the predicted quaternion $\hat{\bm{q}}$ and the ground truth $\bm{q}$,
\textit{i.e.}, $E_q = 2\arccos(\hat{\bm{q}}^T \bm{q} )$.
The translation error is defined as the difference between the predicted value $\hat{\bm{t}}$ and the ground truth $\bm{t}$,
\textit{i.e.}, $E_t = \|\hat{\bm{t}} - \bm{t} \|_2$.
Given an image, the scores for rotation and translation are defined as ${S}_q=E_q$ and ${S}_t=E_t/\|\bm{t}\|_2$, respectively.
The overall score is given as 
\begin{equation}
	S = 
		\begin{cases}
			0, & \text{if}\quad {S}_q < \theta_q \;  \text{and} \;  {S}_t < \theta_t  \\
			{S}_q + {S}_t, & \text{otherwise}
		\end{cases}
\end{equation}
where $\theta_q=0.169^\circ$ and $\theta_t=2.173 \times 10^{-3}$ are the thresholds, 
which are determined by the calibration results of the facility used to create the dataset~\cite{park2021speed+}.

\subsection{Implementation Details}\label{sec:imp}
\textbf{Mesh reconstruction and data preparation.} 
Since the 3D mesh of the satellite is not provided in SPEED+,
we reconstruct the 3D mesh $\mathcal{M}$ and 3D landmarks $\mathcal{P}$ using source samples.
However, due to illumination variations and material discrepancies,
the texture of satellite is not domain-agnostic, 
while the geometrical model are the same for the source and target samples.
Therefore, we reconstruct the texture-less 3D mesh of the satellite to leverage the \textit{geometrical constraints}.

The pipeline for 3D mesh reconstruction and annotation is shown in Fig.~\ref{fig:recon}.
Note that, the source images can have two types of background: the earth background and the black background.
Since the earth background usually introduces noise during reconstruction, 
we first train a classifier to select images with a black background. 
Next, to tackle the scale issue, we select 1000 images using the criterion of 4.5m $\leq \|\bm{t}\|_2 \leq$ 5m, 
where $\bm{t}$ is the translation of the satellite to the camera. 
We first reconstruct the coarse mesh 
using the Multi-View Stereo (MVS) approach provided by colmap~\cite{schoenberger2016sfm, schoenberger2016mvs}. 
Next, we use the Neural Implicit Surfaces (NeuS)~\cite{wang2021neus} approach to refine the mesh.
The refined mesh is then annotated with 5 categories, including antenna 1-3, solar panel, and body.
Meanwhile, we select 11 landmarks on the surface of the mesh, 
following previous works~\cite{chen2019satellite,wang2022revisiting}.
Finally, given the ground-truth poses for source images and the pseudo poses for target images,
we use Blender to render the mesh for fine-grained masks, 
and use the pinhole camera model~\cite{Andrew2006} to obtain 2D keypoints.

\textbf{Architecture details.} 
Our network comprises four modules: a backbone, a mask head, a heatmap head, and a discriminator.
We construct the backbone using a transformer-based HRNet network~\cite{YuanFHLZCW21}, 
\textit{i.e.}, HRFormer-S with 7.8M parameters.
The backbone extracts feature maps at $\frac{1}{4}$ resolution of the input images.
The output channel number is 32.
The heatmap head and the mask head are constructed using an atrous-spatial-pyramid-pooling (ASPP) module~\cite{chen2018encoder}, respectively.
The module consists of five parallel branches: a global average pooling layer, 
a 1 $\times$ 1 convolution layer, and three 3 $\times$ 3 atrous convolution layers with rates of (6, 12, 18). 
Then, all feature maps are concatenated into one feature, whose channel number is adjusted using a 1 $\times$ 1 convolution layer.
The discriminator has $4\times 4$ convolutional layers with channel numbers (16, 32, 64, 128, 1).
The first and the second layers have a stride of 1 while others have a stride of 2.
Each convolutional layer except the last one is followed by a leaky ReLU parameterized by 0.2.

\textbf{Experimental details.} 
We implement the network using the PyTorch library and train our model using the AdamW~\cite{Loshchilov2017FixingWD} optimizer.
All images are resized to the resolution of $640 \times 400$.
The source images are first translated into target-like images using a CycleGAN~\cite{Zhu2017UnpairedIT} to reduce the bias towards the source domain.
During training, we apply different data augmentation strategies on source and target samples using Albumentations~\cite{info11020125} with the default parameter setting.
The data augmentation on target samples composes of random horizontal and vertical flipping, random translating, scaling, and rotating.
On source samples, the additional augmentation includes random gaussian noise and random blur.
We adopt a multi-level learning strategy to enhance domain adaptation by applying the prediction heads after stages 3 and 4 of the backbone.
The balance parameters of the losses in stages 3 and 4 are set to 0.1 and 1, respectively.
During pseudo-label generation, threshold $N_{th}$ is empirically set to 8. 

\subsection{Ablation Study}\label{sec:abl}
We conduct a series of ablation experiments to investigate the critical components of our approach,
including self-training, adversarial training, mask prediction, 
and pseudo-label generation with the \textit{geometrical constraints}.
Due to the unavailability of pose labels of target samples,
we manually annotate 100/50 images from the \texttt{lightbox}/\texttt{sunlamp} domain by selecting semantic keypoints and solving the PnP problem.
We adopt the manually annotated samples as the validation set.
For each setting, we take 3 rounds to optimize Eq.~(\ref{eq:self2}) and define the first round as the pretraining stage.
During pretraining, we set the initial learning rate to 0.001 and train the network for 12 epochs with a batch size of 5.
For the second and the third rounds, 
we decrease the learning rate to 0.0005 and optimize the network for 10 epochs, which contains 10, 000 steps.
The results are reported in Table~\ref{tab:ab1}.
The neural network trained after round $j$ under the $i$th setting is denoted by $\mathcal{G}_{i}^{j}$.
Especially, model $\mathcal{G}^0_1$ is the trivial baseline, 
as it is trained on target-like images transformed using a CycleGAN and no other UDA method is used.

\begin{table*}[!t]
	\centering
	\tabcolsep=0.25cm
	\caption{Ablation Study of the key components on the SPEED+ dataset.}
	\label{tab:ab1}
	\begin{threeparttable}
	\begin{tabular}{@{}lcccccccccc@{}}
		\toprule
		\multirow{2}{*}{Model}& w/o & w/o  & part  & \multirow{2}{*}{round}  & \multicolumn{3}{c}{\texttt{lightbox}} &  \multicolumn{3}{c}{\texttt{sunlamp}} \\
		\cmidrule(lr){6-8} \cmidrule(lr){9-11}
		& 				adv.\tnote{1} & con.\tnote{2} & num &  	& $E_t$ [m]  & $E_q$ [$^\circ$] & $S$ [-] & $E_t$ [m] & $E_r$ [$^\circ$] & $S$ [-] \\
		\midrule
		\multirow{3}{*}{$\mathcal{G}_1$} &   						&\multirow{3}{*}{$\surd$}&	\multirow{3}{*}{0} 	& 0	& 1.0069 & 36.0853 & 0.7952 & 0.9986 & 39.3665 & 0.8651\\
		&						  	& 					&															& 1 & 0.8524 & 21.7186 & 0.5180	& 0.9242 & 30.2367 & 0.6861\\
		&						  	& 					&														    & 2 & 0.9339 & 24.4870 & 0.5691	& 0.9738 & 29.0898 & 0.6846	\\
		\midrule
		\multirow{3}{*}{$\mathcal{G}_2$} &\multirow{3}{*}{$\surd$}&\multirow{3}{*}{$\surd$}&	\multirow{3}{*}{0}  & 0	& 0.2793 & 20.6284 & 0.4251 & 0.5211 & 32.1571 & 0.6576\\
		&						  	& 					&															& 1 & 0.2454 & 15.1444 & 0.3136	& 0.2214 & 16.4786 & 0.3324\\
		&						  	& 					&															& 2 & 0.2307 & 10.8686 & 0.2315	& 0.1536 & 11.2180 & 0.2226\\
		\midrule
		\multirow{3}{*}{$\mathcal{G}_3$} &\multirow{3}{*}{$\surd$}  &\multirow{3}{*}{$\surd$}&\multirow{3}{*}{1}	& 0	& 0.3662 & 20.4000 & 0.4175 & 0.4850 & 21.0174 & 0.4390\\
										 &						 	& 						 &    					& 1	& 0.2928 & 14.6781 & 0.3033	& 0.2820 & 14.6238 & 0.3011\\
										 &						  	& 				  		 &	 					& 2	& 0.3767 & 11.1182 & 0.2523 & 0.1669 & 11.9669 & 0.2359\\
		\midrule
		\multirow{3}{*}{$\mathcal{G}_4$} &\multirow{3}{*}{$\surd$}&	&\multirow{3}{*}{5} & 0 & 0.3235 & 16.6921 & 0.2865	& 0.2043 & 8.9460 &	0.1873\\
										 &						  &	&					& 1 & 0.2342 & 11.1159 & 0.2386	& 0.1310 & 6.3235 &	0.1310\\
										 &   					  &	& 					& 2 & 0.2310 & 11.7523 & 0.2463	& 0.1104 & 5.2222 & 0.1103\\
		\midrule
		\multirow{3}{*}{$\mathcal{G}_5$} &\multirow{3}{*}{$\surd$}&\multirow{3}{*}{$\surd$}	&\multirow{3}{*}{5} & 0 & 0.3235 & 16.6921 & 0.2865 & 0.2043 & 8.9460 & 0.1873\\
										 &						  &							&					& 1 & 0.2346 & 10.8023 & 0.2302	& 0.1068 & 5.2535 & 0.1108\\
										 &   					  &							& 					& 2 & \bfseries 0.2039 & \bfseries 10.5052 & \bfseries 0.2205	& \bfseries 0.0963 & \bfseries 4.6960 & \bfseries 0.0988\\
		\bottomrule
	\end{tabular}
	 \begin{tablenotes}
        \footnotesize
        \item[1] with/without adversarial training;
        \item[2] with/without domain-agnostic \textit{geometrical constraints}.
      \end{tablenotes}
    \end{threeparttable}
\end{table*}

\textbf{Self-training}. 
In each setting, models $\mathcal{G}^{1}$ and $\mathcal{G}^{2}$ achieve smaller pose estimation errors than model $\mathcal{G}^{0}$.
This demonstrates that the self-training framework can promote model performance for the UDA task in satellite pose estimation.

\textbf{Adversarial training}. 
We construct a baseline $\mathcal{G}_1$ within the basic framework by optimizing Eq.~(\ref{eq:self1}).
Then, we extend the framework with adversarial training, by aligning the features extracted by the backbone.
The model trained in the extended framework is denoted by $\mathcal{G}_2$.
For each round, model $\mathcal{G}_2^i$ significantly outperforms the baseline $\mathcal{G}_1^i$, $i=0,1,2$.
After the pretraining stage, 
model $\mathcal{G}_2^0$ reduce the translation error from 1.0069m/0.9986m to 0.2793m/0.5211m for the \texttt{lightbox}/\texttt{sunlamp} domain.
When round = 2, model $\mathcal{G}_2^2$ reduces the scores by half for both domains.
These results demonstrate the effectiveness of adversarial training.

\begin{figure}[t]
	\centering
	\includegraphics[width=\linewidth]{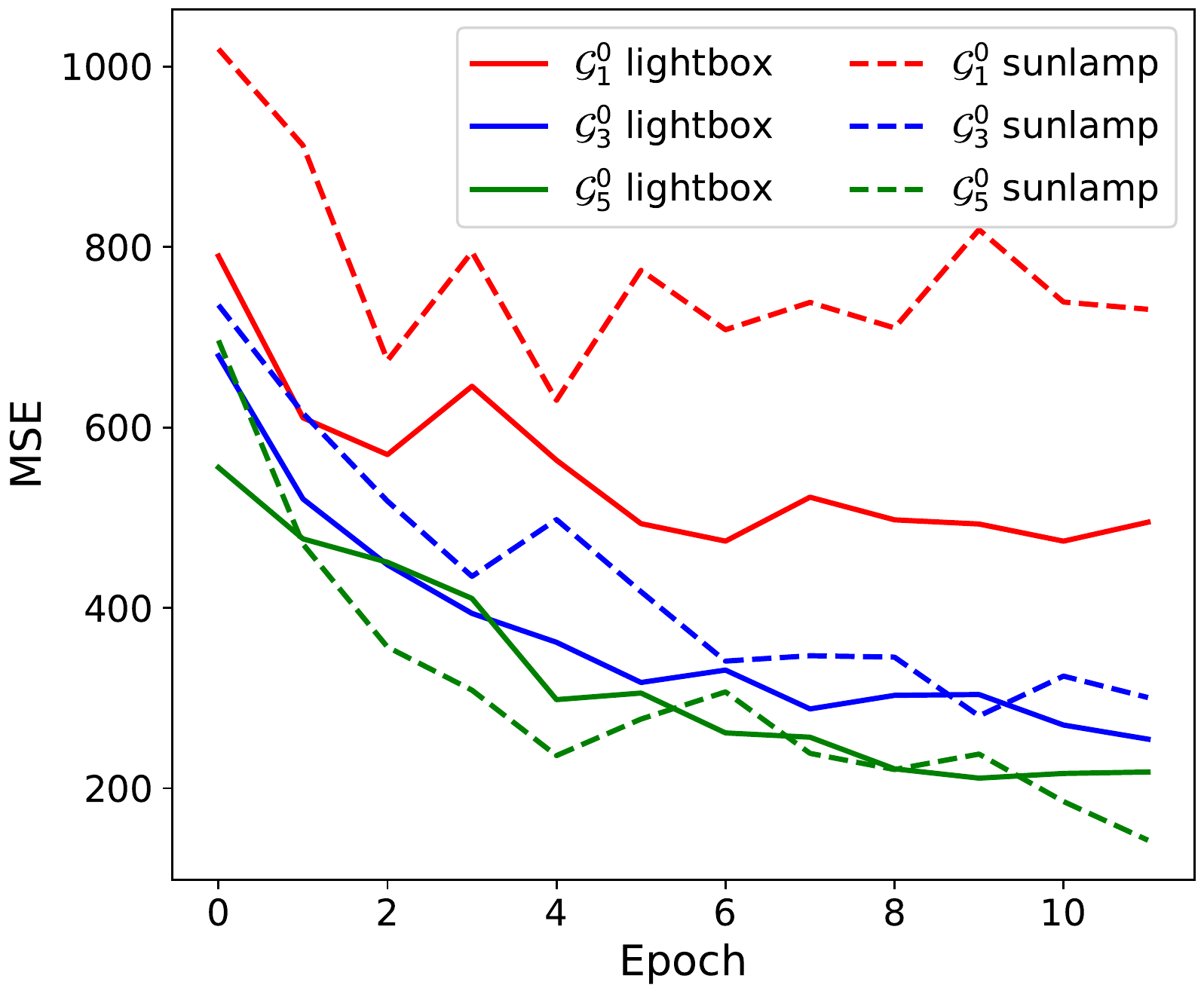}
	\caption{The mean square errors (MSE) between the predicted and ground-truth keypoints in the pretraining stage.}
    \label{fig:nme_epoch}
\end{figure}

\textbf{Mask prediction}. 
We further extend the baseline with multi-task learning by appending a mask head after the backbone.
The mask head predicts a binary mask (in model $\mathcal{G}_3$) 
or a fine-grained mask (in model $\mathcal{G}_5$) of the satellite.
Different from model $\mathcal{G}_2$, we apply adversarial training by aligning the predictions of the mask head.
Model $\mathcal{G}_3$ achieves more accurate pose estimation results 
than model $\mathcal{G}_2$ in all rounds except the last one in terms of total score $S$.
Another experimental evidence is provided by the comparison between 
model $\mathcal{G}_3$ and model $\mathcal{G}_5$.
When round = 0, model $\mathcal{G}_5^0$ significantly reduces 
the estimation error by nearly 30\% and 50\% in terms of score $S$ for both domains.  
The key reason is that the mask head in model $\mathcal{G}_5$ provides more fine-grained predictions and 
thus effectively enhances structural constraints and contextual information.
Moreover, using the proposed self-training framework, 
model $\mathcal{G}_5^2$ achieves the best pose estimation results in all metrics and on all tasks.
Therefore, leveraging fine-grained segmentation as an auxiliary task has positive impacts on 2D keypoints regression, 
resulting in better performance of satellite pose estimate across domains.

\begin{figure}[t]
    \centering
    \subfloat[\texttt{lightbox}]{
    \includegraphics[width=.49\linewidth]{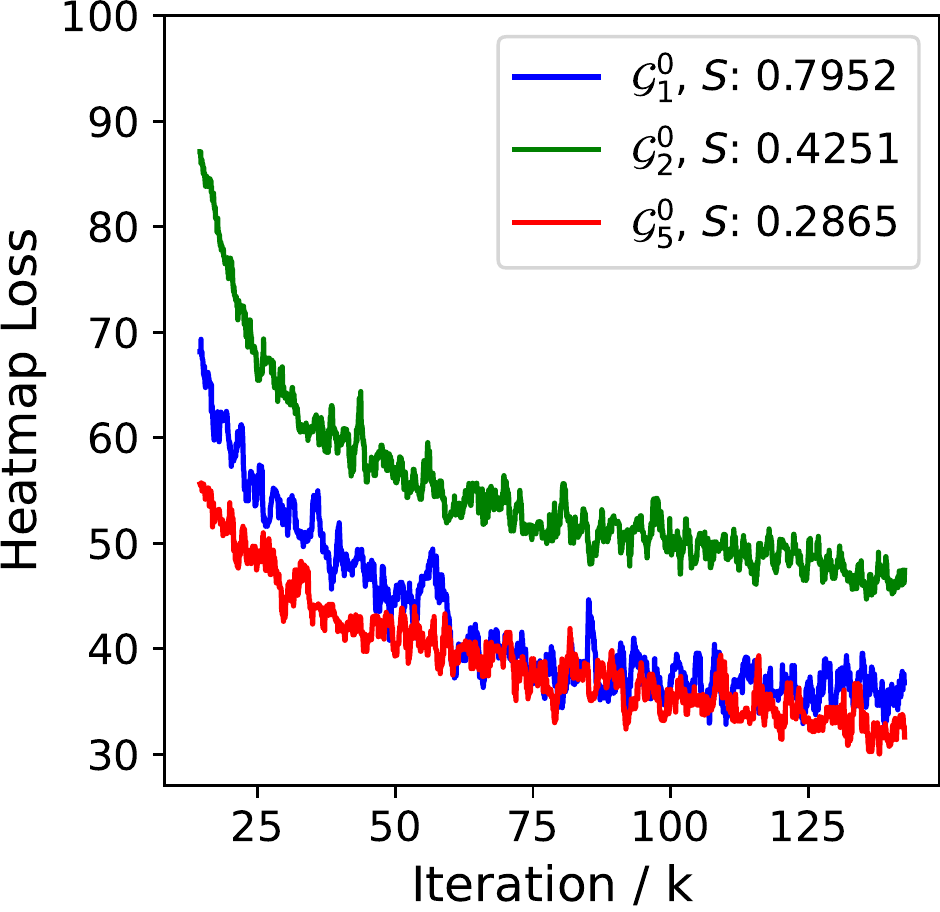}
    }
    \subfloat[\texttt{sunlamp}]{
    \includegraphics[width=.49\linewidth]{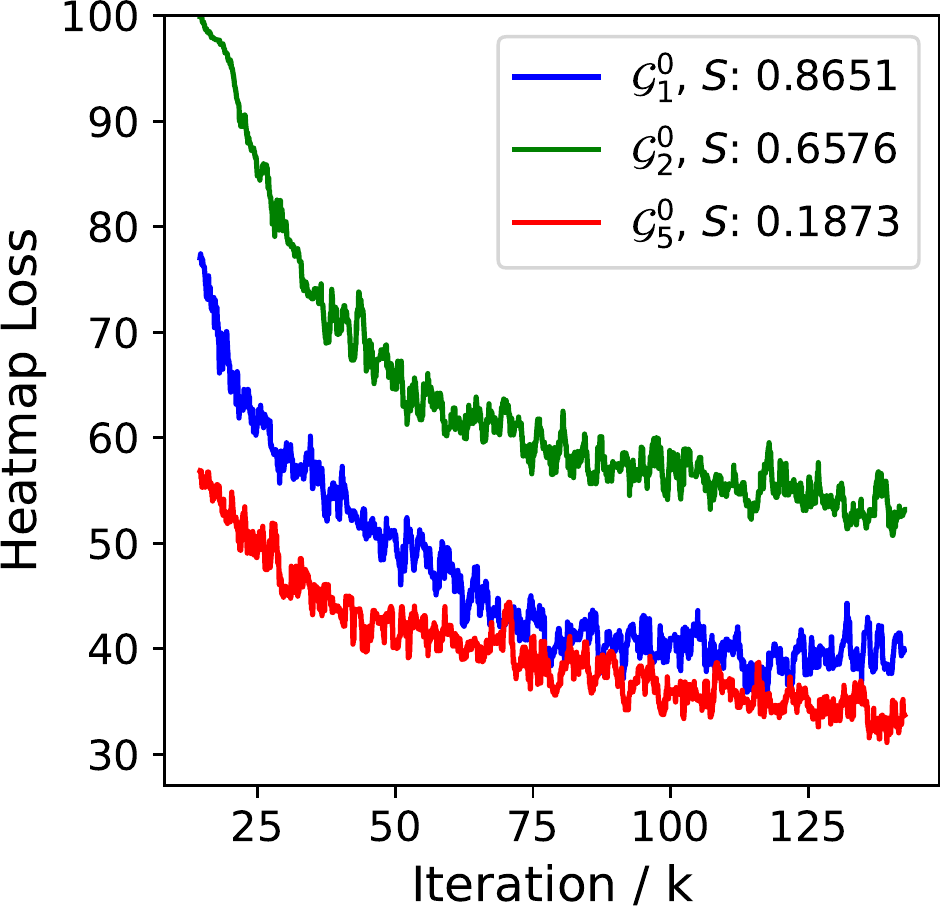}
    }
    \caption{
    Comparisons of the training curves of model $\mathcal{G}_0^0$, $\mathcal{G}_2^0$, and $\mathcal{G}_5^0$.}
    \label{fig:losses}
\end{figure}

To better analyze the function of the fine-grained segmentation, 
we compare the mean square error (MSE) between the ground-truth and the predicted keypoints on the validation set after each epoch.
Specifically, we report the MSEs of the pretrained models, 
including $\mathcal{G}_1^0$, $\mathcal{G}_3^0$, and $\mathcal{G}_5^0$,
and the results are shown in Fig.~\ref{fig:nme_epoch}.
Note that, model $\mathcal{G}_1^0$ always achieves the largest MSEs on \texttt{lightbox} and \texttt{sunlamp} domains,
while $\mathcal{G}_3^0$ benefits from the binary segmentation task.
Moreover, $\mathcal{G}_5^0$ shows the highest accuracy by predicting the fine-grained masks. 
It illustrates that the fine-grained segmentation can effectively improve the domain adaptation performance,
and thus prevent the optimization of Eq.~(\ref{eq:self2}) converging at local optima.

\begin{figure*}
	\centering
    \subfloat[\texttt{lightbox}]{
		\begin{minipage}[b]{.95\linewidth}
		\includegraphics[width=\textwidth]{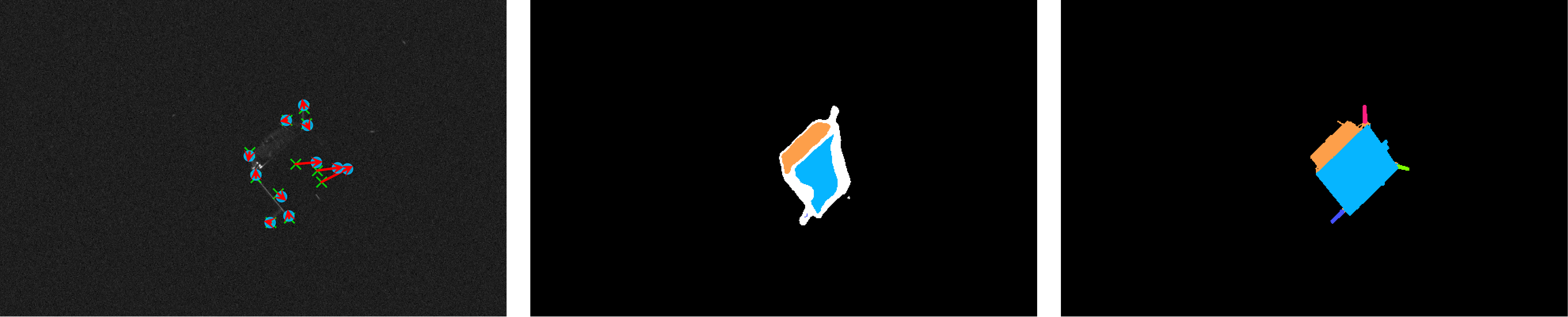}\\\vspace{-9pt}

		\includegraphics[width=\textwidth]{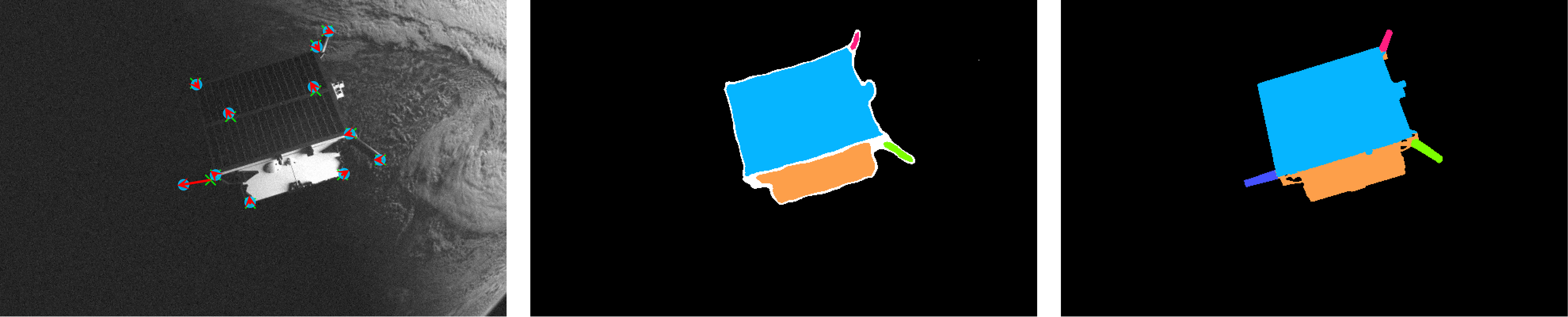}\\\vspace{-9pt}
        
		\includegraphics[width=\textwidth]{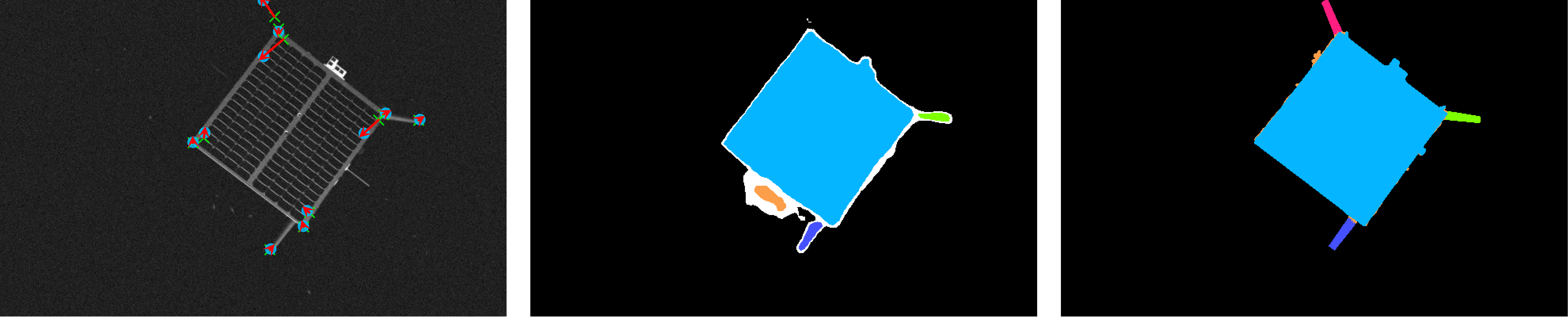}
		\end{minipage}
	}\\\vspace{-3pt}
	\subfloat[\texttt{sunlamp}]{
		\begin{minipage}[b]{.95\linewidth}
			\includegraphics[width=\textwidth]{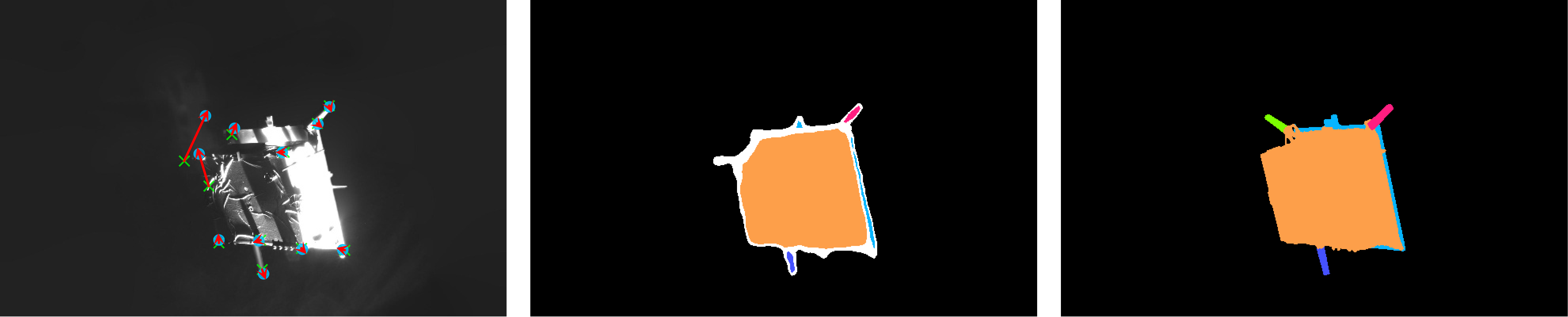}\\\vspace{-9pt}

			\includegraphics[width=\textwidth]{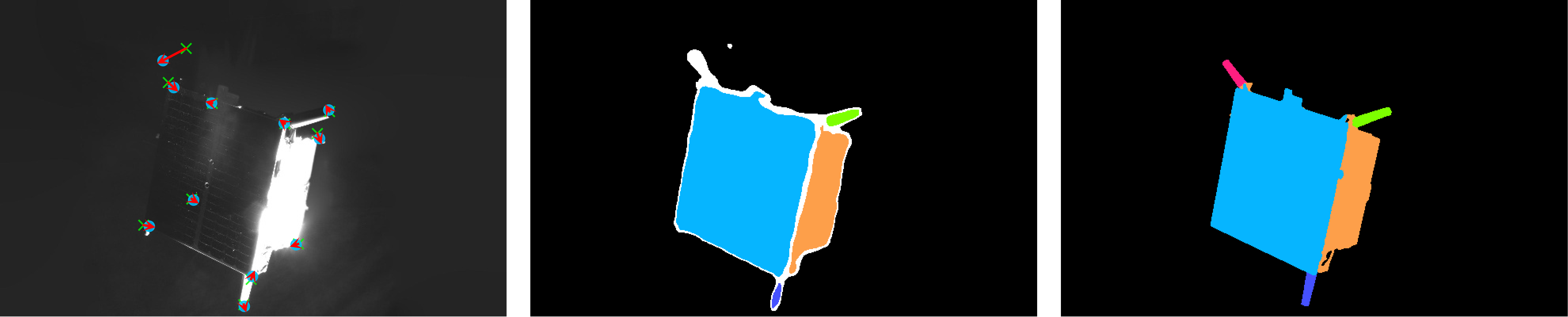}\\\vspace{-9pt}

			\includegraphics[width=\textwidth]{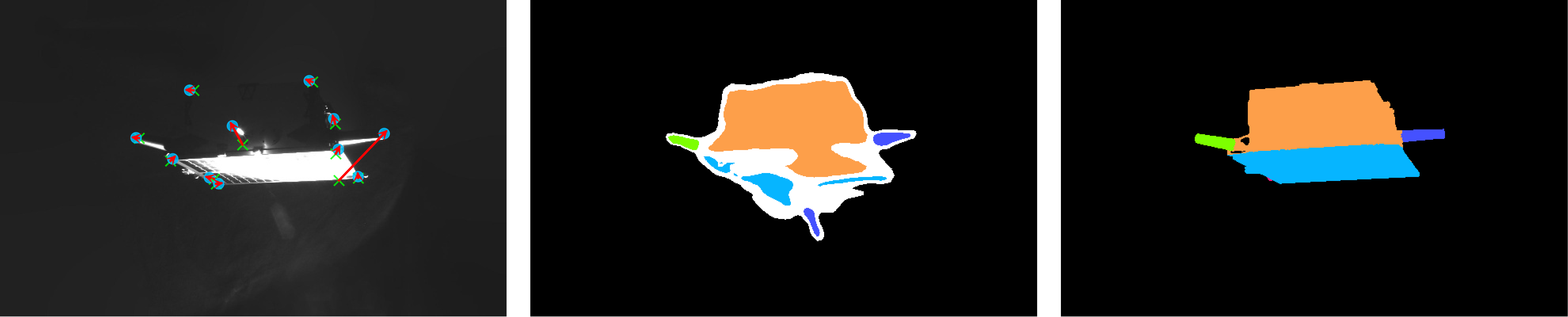}
		\end{minipage}
	}\vspace{-3pt}
	\caption{
		Visualization of pseudo labels of target samples. 
		The first column shows the pseudo keypoints generated without and with the \textit{geometrical constraints} 
        in green crosses and blue points, respectively. 
		The red vectors illustrate the differences between the two types of pseudo keypoints.
		The second/last column shows pseudo masks generated without/with the \textit{geometrical constraints}.
		\label{fig:pseudo}
	}
\end{figure*}

\begin{figure*}[t]
	\centering
    \subfloat[\texttt{lightbox}]{
		\begin{minipage}[b]{\linewidth}
		\includegraphics[width=0.33\textwidth]{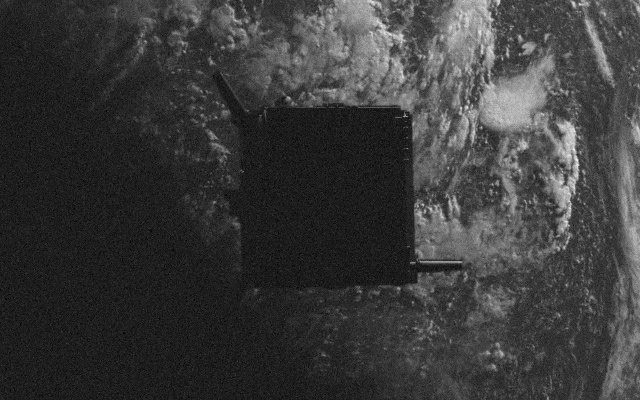}
		\includegraphics[width=0.33\textwidth]{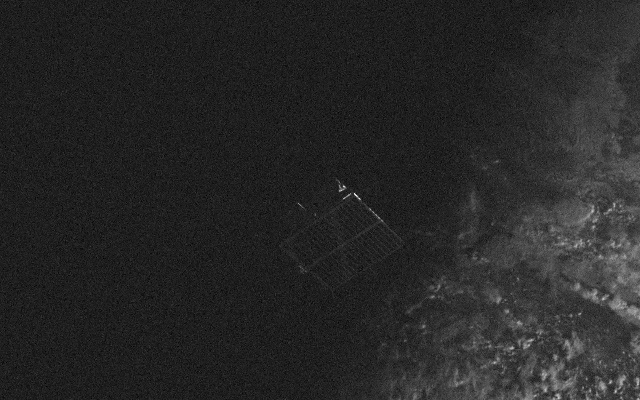}
		\includegraphics[width=0.33\textwidth]{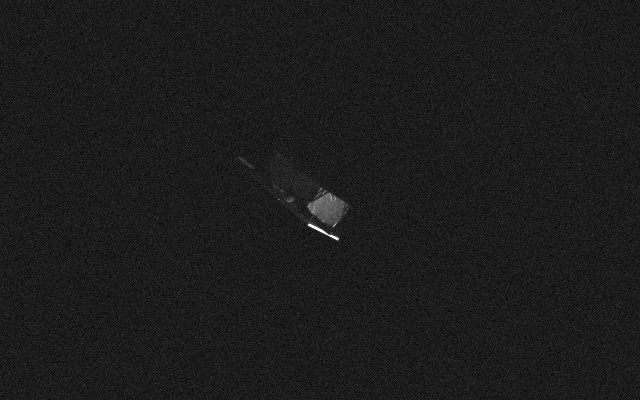}\\

		\includegraphics[width=0.33\textwidth]{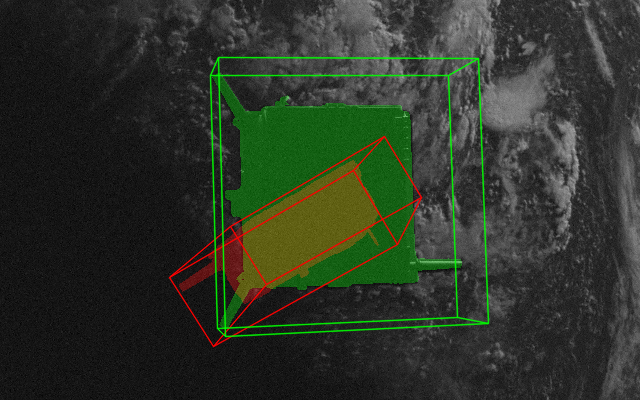}
		\includegraphics[width=0.33\textwidth]{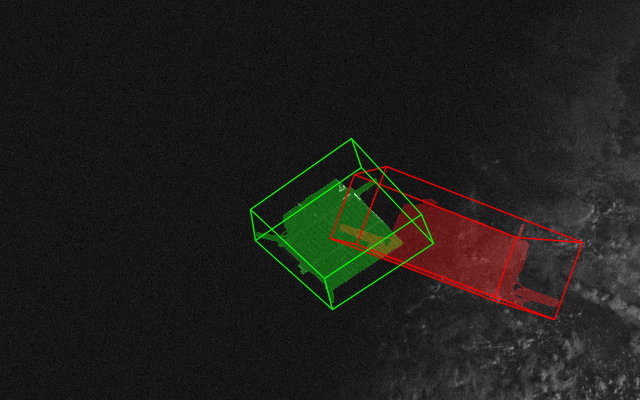}
		\includegraphics[width=0.33\textwidth]{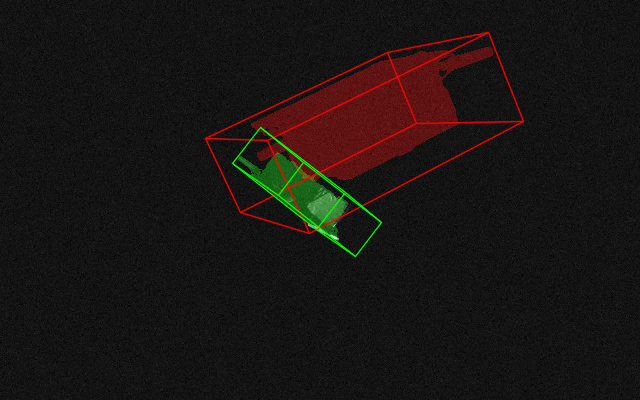}
		\end{minipage}
	}\\
	\subfloat[\texttt{sunlamp}]{
		\begin{minipage}[b]{\linewidth}
			\includegraphics[width=0.33\textwidth]{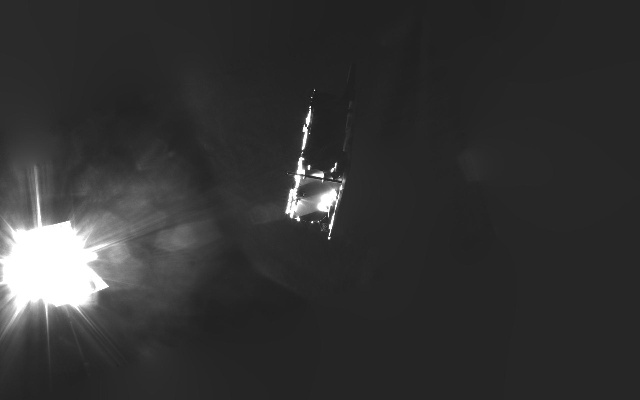}
			\includegraphics[width=0.33\textwidth]{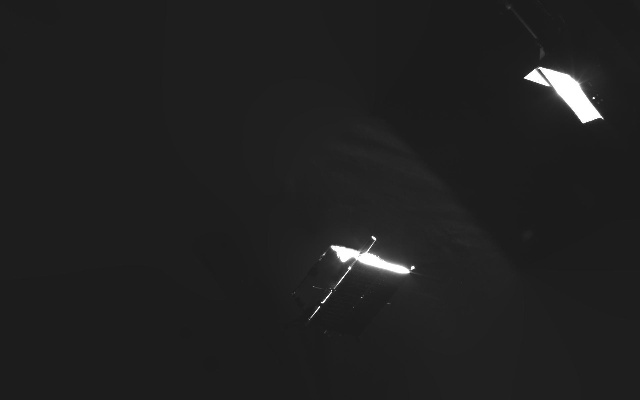}
			\includegraphics[width=0.33\textwidth]{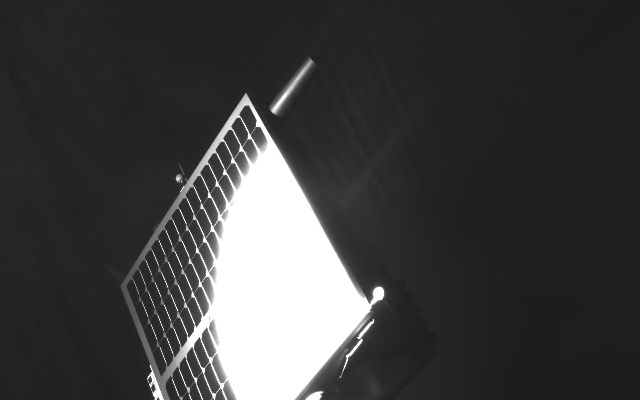}\\

			\includegraphics[width=0.33\textwidth]{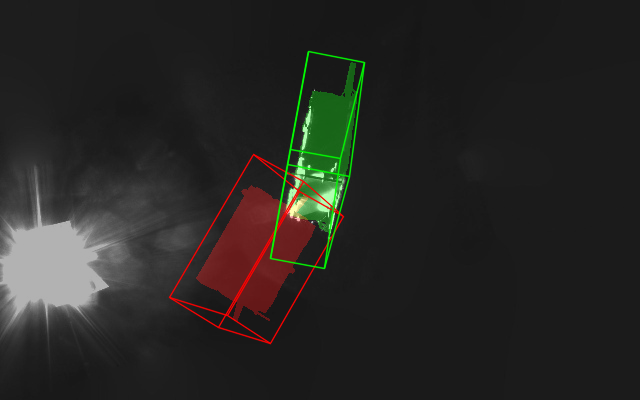}
			\includegraphics[width=0.33\textwidth]{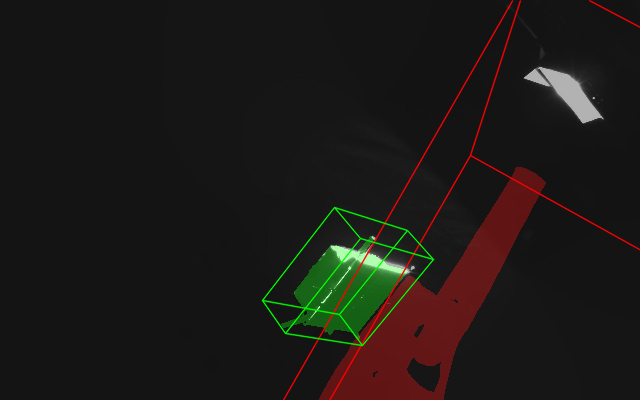}
			\includegraphics[width=0.33\textwidth]{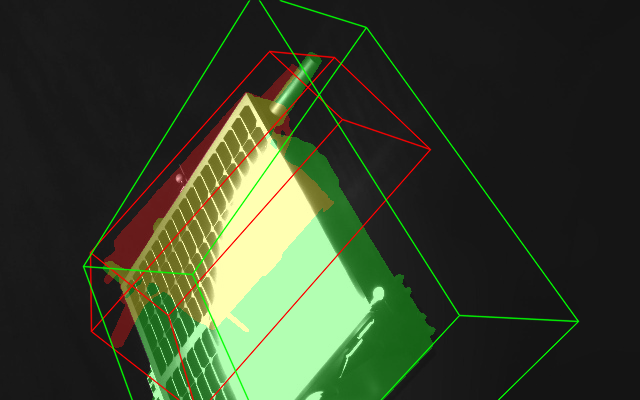}
		\end{minipage}
	}
	\caption{
		Visualization of pose estimation on target samples. 
		The results achieved with and without our framework are shown in green and red colors, respectively.
		\label{fig:wouda}
	}
\end{figure*}

\textbf{Geometrical constraints}. 
We also study the role of the \textit{geometrical constraints}, which are used during pseudo-label generation.
We directly generate pseudo heatmaps and pseudo masks according to model predictions.
The model trained using this setting is denoted by $\mathcal{G}_4$.
In terms of three metrics on both domains, 
models $\mathcal{G}_4^1$ and $\mathcal{G}_4^2$ show degraded performance compared to $\mathcal{G}_5^1$ and $\mathcal{G}_5^2$, respectively.
The performance degradation can be ascribed to the annotation noise in pseudo labels.
With the \textit{geometrical constraints}, 
models $\mathcal{G}_5^1$ and $\mathcal{G}_5^2$ are trained using more clean and more accurate pseudo labels and thus show superior performance.
The visual comparison between pseudo labels generated with and without \textit{geometrical constraints} is illustrated in Fig.~\ref{fig:pseudo}.

\textbf{Multi-task learning.} 
We adopt adversarial training and mask segmentation to promote keypoint heatmap regression.
In Fig.~\ref{fig:losses}, we study the effectiveness of multi-task learning strategies,
by comparing the heatmap losses of model $\mathcal{G}_0^0$, $\mathcal{G}_2^0$, and $\mathcal{G}_5^0$ during training.
On each domain,
model $\mathcal{G}_0^0$ overfits more to source samples and generalizes less to target images,
because it is only trained using source data.
Model $\mathcal{G}_2^0$ performs feature level adaptation using adversarial training.
Although model $\mathcal{G}_2^0$ has larger heatmap losses, 
it achieves better pose estimation accuracy than model $\mathcal{G}_0^0$.
In contrast, $\mathcal{G}_5^0$ performs output level adaptation by introducing fine-grained segmentation,
resulting in the smallest losses and the best pose estimation accuracy on each domain.
The reason is that the feature level adaptation is performed in the high-dimensional space,
leading to the alignment of easier patterns~\cite{Tsai2018LearningTA}.
Consequently, the feature distributions cannot be effectively matched.
This demonstrates the effectiveness of our multi-task learning strategies.

\subsection{Comparison with the State-of-the-Art Methods}\label{sec:comp}
We take KRN~\cite{AAS2019park} and SPNv2~\cite{park2022robust} as the baseline methods. 
SPNv2~\cite{park2022robust} is based on EfficientDet~\cite{Tan2020EfficientDetSA} and comprises three prediction heads: 
the EfficientPose head~\cite{Groos2021EfficientPoseSS} for object presence, bounding box, target rotation and translation;  
the heatmap head for the 2D heatmaps; the segmentation head for the binary mask of the satellite.
Other technologies employed by SPNv2 include multi-scale design~\cite{Tan2020EfficientDetSA,Groos2021EfficientPoseSS},
extensive data augmentation~\cite{Buslaev2020AlbumentationsFA}, 
style augmentation~\cite{Jackson2019StyleAD}, AdaBN~\cite{2016Revisiting}, and entropy minimization.
Different from these methods, 
we utilize the \textit{geometrical constraints} to develop a self-training framework and 
explore the fine-grained segmentation to boost performance.

To achieve better performance,
we add an upsampling layer followed by a $3\times 3$ convolution after the backbone and thus increase the feature resolution by a scaling factor 2.
We take multiple rounds to optimize Eq.~(\ref{eq:self2}) and then compare our method with 
KRN~\cite{AAS2019park}, SPNv2~\cite{park2022robust}, and top-performing methods in SPEC2021.
The results are listed in Table~\ref{tab:final_results_offline}.
On the \texttt{sunlamp} domain, our approach outperforms all other methods in terms of translation and rotation scores,
taking the 1st place in the challenge.
More importantly, our method surpasses KRN~\cite{AAS2019park} trained using real annotations by more than half in terms of score $S$.
On the \texttt{lightbox} domain,
our approach shows competitive performance and has won the 3rd place in the challenge.
Figure~\ref{fig:wouda} visualizes results estimated by the models trained with and without the proposed framework.
Note that, our model can handle intense imaging noise, complicated illuminations, surface reflection, and pose variations.

\begin{table*}[t]
	\caption{Comparison with the state-of-the-art methods and the top-performing methods in SPEC2021.}
	\label{tab:final_results_offline}
	\centering
	\tabcolsep=0.3cm
	\begin{threeparttable}
	\begin{tabular}{@{}lcccccc@{}}
		\toprule
		\multirow{2}{*}{Methods/Team} & \multicolumn{3}{c}{\texttt{lightbox}} &  \multicolumn{3}{c}{\texttt{sunlamp}}  \\
		\cmidrule(lr){2-4} \cmidrule(lr){5-7}
		& $S_t$	 & $S_q$ & $S$ & $S_t$  & $S_q$ & $S$ \\
		\midrule
		KRN \cite{AAS2019park} 		 & -  & - & 1.12 & - & - &  3.73  \\
		\quad + Style aug. \cite{Jackson2019StyleAD} \tnote{1} & -  & - & 0.81 & - & - &  1.32\\
		\quad + Oracle \tnote{2}     & -  & - & 0.15 & - & - &  0.13 \\
		SPNv2 \cite{park2022robust}  & -  & - & 0.122 & - & - &  0.198 \\
		\midrule
		UT Austin Seeker RD   		 & 0.3626  & 0.1532	& 0.5158 & 0.3589  & 0.1227	& 0.4816 \\
		bbnc						 & 0.0940  & 0.4344	& 0.5284 & 0.0794  & 0.3832	& 0.4626\\
		chusunhao					 & 0.0303  & 0.2859	& 0.3162 & 0.0573  & 0.7567	& 0.8140 \\
		u3s\_lab 			 		 & 0.0530  & 0.1692	& 0.2221 & 0.0297  & 0.1089	& 0.1386 \\
		haoranhuang\_njust			 & 0.0279  & 0.1419	& 0.1698 & 0.0248  & 0.1467	& 0.1715 \\
		VPU					 		 & 0.0178  & 0.0799	& 0.0977 & 0.0072  & 0.0493 & 0.0565 \\
		TangoUnchained 				 & \bfseries 0.0139  & \bfseries 0.0556 & \bfseries 0.0695 & 0.0109  & 0.0750 & 0.0859 \\
		lava1302 (ours)	   		     & 0.0464  & 0.1163	& 0.1627 & \bfseries 0.0069  & \bfseries 0.0476 &\bfseries  0.0545 \\
		\bottomrule
	\end{tabular}
	 \begin{tablenotes}
        \footnotesize
        \item[1] trained with style augmentation;
        \item[2] trained with the ground-truth poses of target samples.
      \end{tablenotes}
    \end{threeparttable}
\end{table*}

\begin{table*}[t]
	\caption{Run-time analysis with different backbones and feature resolutions.}
	\label{tab:time}
	\centering
	\tabcolsep=0.2cm
	\begin{tabular}{lccccccccccc}
		\toprule
		\multirow{2}{*}{Backbone} &{Feat.} & {Param.} & Mem. & {Time}  & \multicolumn{3}{c}{\texttt{lightbox}} &  \multicolumn{3}{c}{\texttt{sunlamp}}\\
		\cmidrule(lr){6-8} \cmidrule(lr){9-11}
		& {Stride} & 	[M]				    &  [G]& 			[ms]			& $E_\textrm{t}$ [m]		 & $E_\textrm{r}$ [rad] & $S_\textrm{pose}$ [-] & $E_\textrm{t}$ [m] & $E_\textrm{r}$ [rad] & $S_\textrm{pose}$ [-] \\
		\midrule
		HRNet-W32 \cite{WangSCJDZLMTWLX19} & 4 & 28.01& 1.69 & \bfseries 36.0 &  0.218   & 12.259 & 0.253 & 0.119  &	8.259 & 0.163 \\
		HRFormer-S \cite{YuanFHLZCW21}     & 4 & \bfseries 8.19 & \bfseries 1.53 & 57.1 &	0.204   & 10.505 & 0.221 & 0.096  &	4.696 & 0.099 \\
		HRFormer-S \cite{YuanFHLZCW21}     & 2 & 8.21 & 1.84 & 57.3 & \bfseries 0.138	& \bfseries 3.223 &\bfseries 0.078 &\bfseries 0.042  &\bfseries	1.616 &\bfseries 0.036 \\
		\bottomrule
	\end{tabular}
\end{table*}

\subsection{Runtime Performance}\label{sec:runtime}
We follow the experimental settings in ablation study and 
compare the runtime performance of the models using different backbones and feature map resolutions.
The experiments are conducted on a PC with an Nvidia GTX 3090 GPU.
Table~\ref{tab:time} reports the results.
We first replace the transformer-based backbone with a CNN-based backbone, \textit{i.e.}, HRNet~\cite{WangSCJDZLMTWLX19}.
Although the running time is shorter at the expense of larger parameter sizes and memory consumption,
the pose estimation performance drops from 0.221/0.099 to 0.253/0.163 on \texttt{lightbox}/\texttt{sunlamp} in terms of score $S$.
Furthermore, we add an upsampling layer followed by a $3\times 3$ convolution after the backbone, 
to increase the feature resolution by a scaling factor 2.
We observe significant improvements of pose estimation at the expense of increased computation.
However, the increase in running time is only about 0.35\% (which is negligible)  
since the upsampling module and the prediction heads are very simple.

\section{LIMITATION AND CONCLUSION}
\subsection{Limitation}
One apparent limitation of our method is that only the predicted heatmaps are used to generate pseudo poses in Sec.~\ref{sec:solution}, 
while the predicted masks are ignored.
Another limitation would be the sparse representation of the satellite using a set of keypoints.
Note that the pose estimation could be less stable when these keypoints are invisible due to truncation, low light, or high reflection.
The third limitation of this work is that the implementations of the projection function $\mathcal{F}$ and 
the rendering function $\mathcal{R}$ are non-differentiable and performed offline.
Future work is needed, specifically in incorporating a differentiable rendering engine 
to simultaneously achieve online self-training and pose refinement based on fine-grained masks.

\subsection{Conclusion}
This paper explores the domain-agnostic \textit{geometrical constraints} 
to achieve unsupervised domain adaptation in satellite pose estimation.
The task is formulated as a minimization problem in a self-training framework 
by taking the target poses as latent variables.
Meanwhile, the fine-grained segmentation is introduced as an auxiliary task to improve performance.
The experimental results demonstrate that our method achieves superior performance.

\section*{Acknowledgment}
This work was supported by the National Natural Science Foundation of China under Grant 61801491, Grant 61972435, and Grant U20A20185.

\ifCLASSOPTIONcaptionsoff
  \newpage
\fi



\bibliographystyle{IEEEtran}
\bibliography{IEEEabrv, bare_jrnl}




%



\begin{IEEEbiography}[{\includegraphics[width=1in,height=1.25in,clip,keepaspectratio]{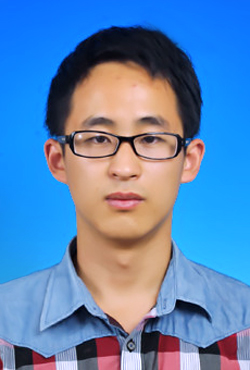}}]{Zi Wang}
	received the B.E. degree in measuring and controlling technologies and instruments from Tianjin University (TJU), Tianjin, China, in 2016, 
	and the M.E. degree in Aeronautical and Astronautical Science and Technology from National University of Defense Technology (NUDT), Changsha, China, in 2018. 
	He was a visiting Ph.D. student in Sun Yat-sen University from 2021 to 2022.
	He is currently pursuing the Ph.D. degree with the College of Aerospace Science and Engineering, NUDT. 
	His current research interests include transfer learning and object pose estimation.
\end{IEEEbiography}

\begin{IEEEbiography}[{\includegraphics[width=1in,height=1.25in,clip,keepaspectratio]{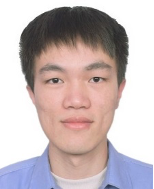}}]{Minglin Chen} 
        is currently a Ph.D. student at Sun Yat-Sen University (SYSU). 
        He received his M.Eng. degree in software engineering at the University of Chinese Academy of Science (UCAS) in 2020, and his B.Eng. degree in communication engineering at South China Normal University (SCNU) in 2017. 
        His research interests lie in the intersection of computer vision and computer graphics.
\end{IEEEbiography}

\begin{IEEEbiography}[{\includegraphics[width=1in,height=1.25in,clip,keepaspectratio]{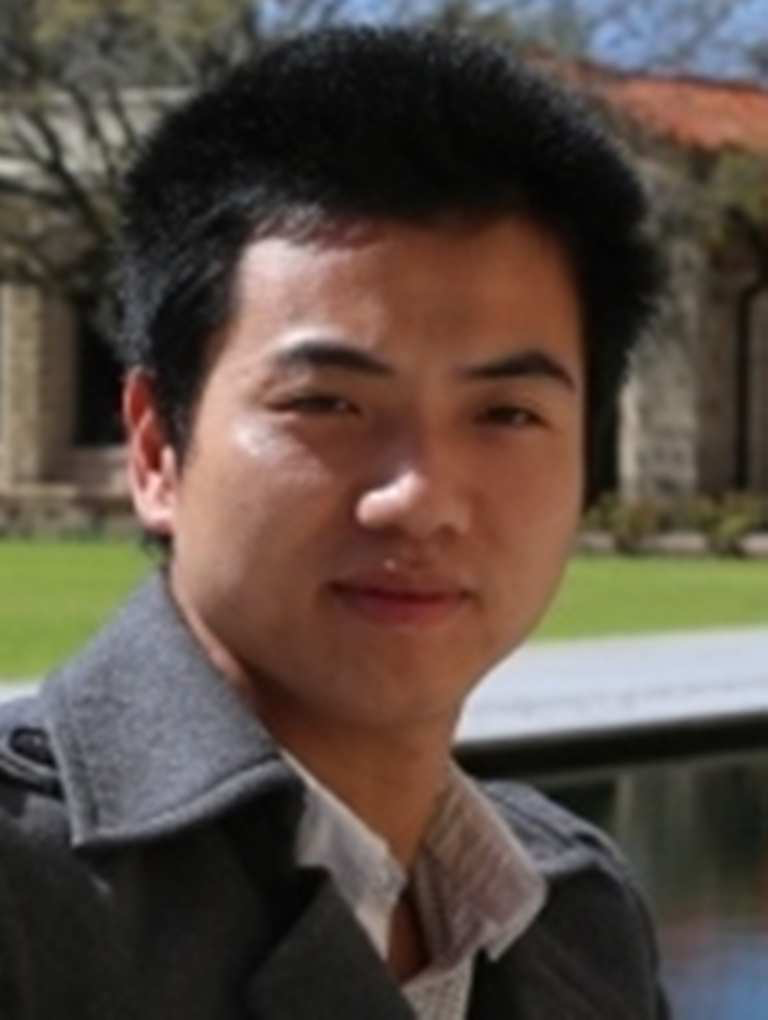}}]{Yulan Guo} 
    received the B.Eng. and Ph.D. degrees from National University of Defense Technology (NUDT) in 2008 and 2015, respectively. 
	He has authored over 100 articles in journals and conferences. 
	His current research interests focus on 3D vision, particularly on 3D feature learning, 3D modeling, 3D object recognition, and scene understanding. 
	He serves as an associate editor for IEEE Transactions on Image Processing, IET Computer Vision, IET Image Processing, and Computers \& Graphics. 
	He also served as a guest editor for IEEE Transactions on Pattern Analysis and Machine Intelligence, and an area chair for CVPR 2023/2021, ICCV 2021, and ACM Multimedia 2021.
\end{IEEEbiography}

\begin{IEEEbiography}[{\includegraphics[width=1in,height=1.25in,clip,keepaspectratio]{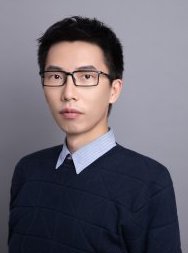}}]{Zhang Li}
	received the Ph.D. degree in biomedical engineering from the Delft University of Technology, Delft, The Netherlands, in 2015.
	He is an Associate Professor with the College of Aerospace Science and Engineering, National University of Defense Technology, Changsha, China. 
	He authored over 20 papers in high-ranking journals and conferences, such as IEEE TRANSACTIONS ON MEDICAL IMAGING and IEEE TRANSACTIONS ON BIOMEDICAL ENGINEERING. 
	His research interests include biomedical image analysis and particularly image registration.
\end{IEEEbiography}

\begin{IEEEbiography}[{\includegraphics[width=1in,height=1.25in,clip,keepaspectratio]{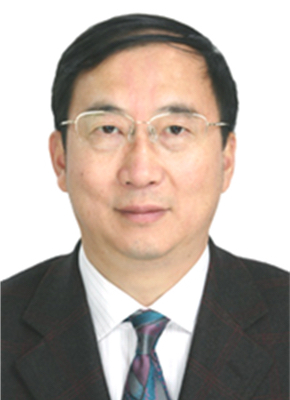}}]{Qifeng Yu}
	received the B.S. degree from Northwestern Polytechnic University, Xi’an, China, in 1981, 
	the M.S. degree from the National University of Defense Technology, Changsha, China, in 1984, 
	and the Ph.D. degree from Bremen University, Bremen, Germany, in 1996.
	He is a Professor with the National University of Defense Technology. 
	He has authored three books and published over 100 articles. 
	His main research fields include image measurement, vision navigation, and close-range photogrammetry.
	Dr. Yu is a member of the Chinese Academy of Sciences.    
\end{IEEEbiography}
\balance




\end{document}